\documentclass[lettersize,journal]{IEEEtran}
\usepackage{amsmath,amsfonts}
\usepackage{algorithmic}

\usepackage{algorithm}
\usepackage{array}
\usepackage[caption=false,font=normalsize,labelfont=sf,textfont=sf]{subfig} 
\usepackage{textcomp}
\usepackage{booktabs}
\usepackage{makecell} 
\usepackage{multirow}
\usepackage{csquotes} 
\usepackage{stfloats}
\usepackage{utfsym}
\usepackage{url}
\usepackage{verbatim}
\usepackage{graphicx}
\graphicspath{{./fig/}}
\usepackage{url}
\usepackage{cite}

\makeatother
\usepackage{hyperref} 
\usepackage{algorithm}
\usepackage{amsmath}
\usepackage{amssymb}
\usepackage{algorithmic}

\makeatletter
\newcommand{\removelatexerror}{\let\@latex@error\@gobble}
\makeatother
\usepackage{cite}
\usepackage{amsmath,amssymb,amsfonts}
\usepackage{algorithmic}
\usepackage{graphicx}
\usepackage{textcomp}
\usepackage{amsmath,amsfonts}
\usepackage{algorithmic}

\usepackage{algorithm}
\usepackage{array}
\usepackage[caption=false,font=normalsize,labelfont=sf,textfont=sf]{subfig} 
\usepackage{textcomp}
\usepackage{booktabs}
\usepackage{makecell} 
\usepackage{multirow}
\usepackage{csquotes} 
\usepackage{stfloats}
\usepackage{utfsym}
\usepackage{url}
\usepackage{verbatim}
\usepackage{graphicx}
\graphicspath{{./fig/}}
\usepackage{url}
\usepackage{cite}
\makeatother
\usepackage{hyperref} 
\usepackage{algorithm}
\usepackage{amsmath}
\usepackage{amssymb}
\usepackage{algorithmic}

\usepackage{tikz}
\usepackage{pifont}
\hypersetup{
colorlinks=true,
linkcolor=red,
citecolor=green,
urlcolor = black
} 
\usepackage{amsmath,amsfonts}
\usepackage{algorithmic}
\usepackage{algorithm}
\usepackage{array}
\usepackage[caption=false,font=normalsize,labelfont=sf,textfont=sf]{subfig}
\usepackage{textcomp}
\usepackage{stfloats}
\usepackage{url}
\usepackage{verbatim}
\usepackage{graphicx}
\usepackage{cite}
\usepackage{amsmath,amsfonts}
\usepackage{algorithmic}

\usepackage{algorithm}
\usepackage{array}
\usepackage[caption=false,font=normalsize,labelfont=sf,textfont=sf]{subfig} 
\usepackage{textcomp}
\usepackage{booktabs}
\usepackage{makecell} 
\usepackage{multirow}
\usepackage{csquotes} 
\usepackage{stfloats}
\usepackage{utfsym}
\usepackage{url}
\usepackage{verbatim}
\usepackage{graphicx}
\graphicspath{{./fig/}}
\usepackage{url}
\usepackage{cite}
\makeatother
\usepackage{hyperref} 
\usepackage{algorithm}
\usepackage{amsmath}
\usepackage{amssymb}
\usepackage{algorithmic}



\usepackage{tikz}
\usepackage{pifont}
\hypersetup{
colorlinks=true,
linkcolor=red,
citecolor=green,
urlcolor = black
} 

\usepackage{booktabs} 
\usepackage{threeparttable}
\usepackage{xcolor}

\newcommand{\best}[1]{\textcolor{blue!95!black}{\textbf{#1}}}

\newcommand{\sm}[1]{\fontsize{#1pt}{18pt}\selectfont}

\usepackage{tikz}
\usepackage{pifont}
\hypersetup{
colorlinks=true,
linkcolor=red,
citecolor=green,
urlcolor = black
} 
\begin{document}

\title{Diffusion-Guided Mask-Consistent Paired Mixing for Endoscopic Image Segmentation}

\author{Pengyu Jie, Wanquan Liu,~\IEEEmembership{Senior Member,~IEEE}, Rui He, Yihui Wen, Deyu Meng,~\IEEEmembership{Senior Member,~IEEE}, Chenqiang Gao
\thanks{Pengyu Jie, Wanquan Liu, and Chenqiang Gao are with the School of Intelligent Engineering, Sun Yat-sen University (Shenzhen Campus), Shenzhen 518107, China. Rui He and Yihui Wen are with the Department of Otolaryngology, The First Affiliated Hospital of Sun Yat-sen University, Guangzhou 510000, China. Deyu Meng is with the School of Mathematics and Statistics, Xi'an Jiaotong University, Xi'an 710049, China (corresponding authors: Wanquan Liu and Chenqiang Gao; emails: liuwq63@mail.sysu.edu.cn; gaochq6@mail.sysu.edu.cn).}}

\markboth{Journal of \LaTeX\ Class Files,~Vol.~14, No.~8, August~2021}%
{Pengyu Jie \MakeLowercase{\textit{et al.}}: Diffusion-Guided Mask-Consistent Paired Mixing for Endoscopic Image Segmentation}

\IEEEpubid{0000--0000/00\$00.00~\copyright~2021 IEEE}
 
\maketitle

\begin{abstract}
Augmentation for dense prediction typically relies on either sample mixing or generative synthesis. Mixing improves robustness but misaligned masks yield soft label ambiguity. Diffusion synthesis increases apparent diversity but, when trained as common samples, overlooks the structural benefit of mask conditioning and introduces synthetic-real domain shift. We propose a paired, diffusion-guided paradigm that fuses the strengths of both. For each real image, a synthetic counterpart is generated under the same mask and the pair is used as a controllable input for Mask-Consistent Paired Mixing (MCPMix), which mixes only image appearance while supervision always uses the original hard mask. This produces a continuous family of intermediate samples that smoothly bridges synthetic and real appearances under shared geometry, enlarging diversity without compromising pixel-level semantics. To keep learning aligned with real data, Real-Anchored Learnable Annealing (RLA) adaptively adjusts the mixing strength and the loss weight of mixed samples over training, gradually re-anchoring optimization to real data and mitigating distributional bias. Across Kvasir-SEG, PICCOLO, CVC-ClinicDB, a private NPC-LES cohort, and ISIC 2017, the approach achieves state-of-the-art segmentation performance and consistent gains over baselines. The results show that combining label-preserving mixing with diffusion-driven diversity, together with adaptive re-anchoring, yields robust and generalizable endoscopic segmentation.

\end{abstract}

\begin{IEEEkeywords}
endoscopic image, conditional diffusion, mixing augmentation, lesion segmentation.
\end{IEEEkeywords}
 
\section{Introduction}
\IEEEPARstart{E}{ndoscopic} imaging is clinically critical because nasal endoscopy, gastroscopy, and colonoscopy are the only routine means of direct mucosal inspection and thus guide lesion detection, biopsy, and surgical intervention. Even small pixel-level segmentation errors, such as blurred or shifted boundaries, can alter diagnostic conclusions and treatment plans~\cite{Tiwari2025}. Deep learning has markedly improved segmentation performance~\cite{point_WangHong_MIA,jiepy_TCSVT,jiepy_eClinicalMedicine}, but these gains rely on large volumes of diverse images and high-quality, pixel-level annotations~\cite{Zhang2025}. In clinical endoscopy, obtaining such annotations requires expert effort and is costly. Data augmentation offers a low-cost way to expand training diversity. 

\begin{figure}[t]
    \vspace{-0.10cm}
	\centering
	\includegraphics[width=\columnwidth]{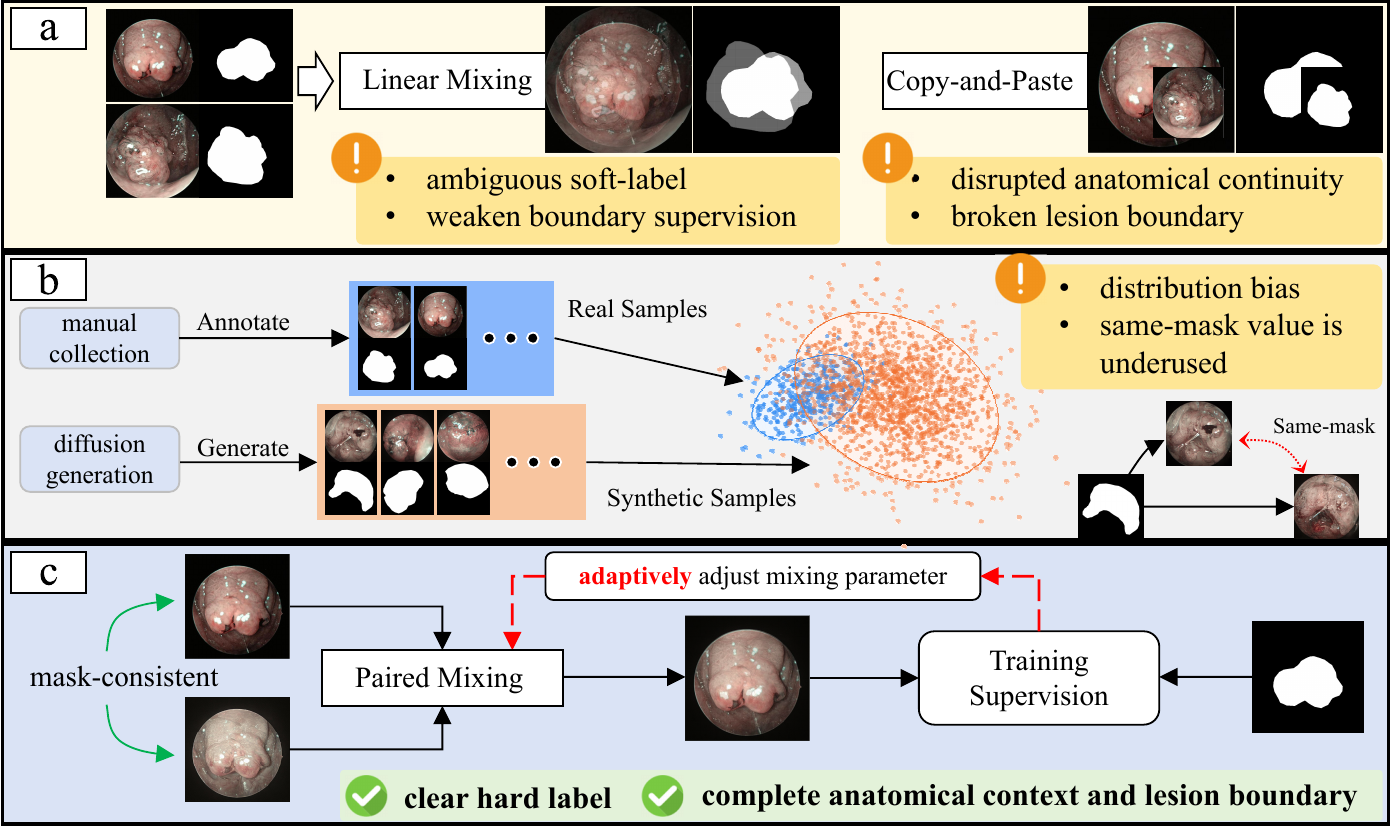}
	\caption{Comparison of augmentation strategies. (a) Inter-sample mixing via linear mixing or copy-and-paste introduces semantic ambiguities and disrupted anatomical continuity. (b) Generative augmentation adds diversity, while underuses same-mask, and induces domain shift. (c) We mix each real image with a mask-consistent synthetic counterpart via diffusion model to preserve lesion geometry and train with hard label supervision, thereby boosting segmentation accuracy and robustness.
    }
	\label{fig:introduction_figure}  
    \vspace{-0.6 cm}
\end{figure}  
\IEEEpubidadjcol

Most augmentation methods for dense prediction fall into two families. The first is sample mixing e.g., Mixup-style linear blending\cite{mixup_2018_ICLR} or Copy-and-paste mixing~\cite{CutMix,PuzzleMix,HSMix}. These methods improve performance by synthesizing intermediate examples between training samples, thereby producing harder composite inputs for the model to learn from.
But for segmentation, they carry a structural problem, i.e., masks from different images almost never align spatially. Standard linear mixing then produces non-binary soft labels (\enquote{half-lesion}) near the boundary (Fig.~\ref{fig:introduction_figure}(a)), where supervision should be explicit~\cite{HOU2024110089,NEURIPS2022_e6f32e64,ClassMix}.
Copy-and-paste mixing methods avoid soft labels, but introduce other issues, i.e., pasted regions may break anatomical integrity by covering or fragmenting lesions, while ignoring foreground--background coupling~\cite{PuzzleMix} (Fig.~\ref{fig:introduction_figure}(a)).
In practice, these methods increase visual diversity at the cost of semantic consistency. This may be acceptable for classification but is unsuitable for dense medical segmentation, because it compromises supervision validity and weakens pixel-level labels, especially near lesion boundaries.

The second family is generative augmentation. Recent work uses diffusion models to synthesize additional endoscopic images and then simply appends these synthetic images and their corresponding masks to the training set as common samples~\cite{konz2024segguided,ControlPolypNet,DiffuseMix,SatSynth}. This approach makes the data look larger and more diverse, but it is still not enough for dense prediction tasks. 
First, it overlooks the key value of mask-conditioned diffusion, the ability to produce multiple distinct appearances for the same semantic mask. That structural property, the same mask with new texture, is not being exploited in current usage (Fig.~\ref{fig:introduction_figure}(b)). 
Second, it assumes synthetic data is harmless. In reality, large-scale inclusion of diffusion-generated images introduces a domain shift~\cite{Domain_Gap_2024_Nature,domain_gap_doge}, since diffusion outputs carry characteristic color, lighting, and microtexture biases. Heavy reliance on such data can push the network toward a synthetic biased distribution~\cite{barbano2025steerable,Domain_Gap_2025_CVPR} and away from real clinical data (Fig.~\ref{fig:introduction_figure}(b)).

To address the limits of mixing in segmentation and the weak use of mask-conditioned synthesis in current generative augmentation, we propose a new method, Mask-Consistent Paired Mixing (MCPMix), that combines both. For each real image-mask pair, we use a conditional generator to create a synthetic image that shares the same mask. We then mix the real and synthetic images to form intermediate samples (Fig.~\ref{fig:introduction_figure}(c)). Mixing happens only in the image appearance space, while supervision always uses the original ground-truth mask (hard label). This paired mixing adds more than variety: it builds a smooth set of samples between synthetic and real appearances under the same geometry, giving the training distribution a clear bridge from synthetic to real.
To address the distribution shift introduced by generative augmentation, we propose an adaptive re-anchoring strategy, Real-Anchored Learnable Annealing (RLA), which adjusts the contribution of synthetic data during training. In early epochs, synthetic images have greater influence on learning. Later, the model adaptively anneals the mix toward real images and then converges to the clinical distribution without hand-tuned schedules. Overall, the framework differs from both conventional mixing and from using generated images as stand-alone training samples. It enables label-preserving mixing by pairing each real image with a mask-consistent, conditionally generated counterpart; it uses appearance diversity without soft labels that weaken pixel-level supervision; and it provides a clear path back to the real clinical domain, reducing drift from synthetic to real.
Our contributions are as follows:
\begin{itemize}
  \item We present a simple yet effective endoscopic augmentation method. We unify the appearance diversity of diffusion-based generation with the interpolative diversity of mixing and use a learnable schedule to control the synthetic ratio, expanding appearance coverage while preserving lesion geometry and improving generalization and robustness.
  \item We propose the MCPMix. To the best of our knowledge, this is the first mixing-based method to apply same-mask real-synthetic pairing for segmentation, preserving target geometry and reducing partial-pixel ambiguity.
  \item We design a dynamic mixing weighting schedule, RLA, for the training process on MCPMix. RLA learns a real-anchored schedule that adaptively down-weights mixed samples and converge training on real images. It is fully differentiable and trained end-to-end.
  \item Extensive experiments across multiple datasets demonstrate clear superiority over state-of-the-art methods in both endoscopic and dermoscopic lesion segmentation, with boundary metrics further confirming consistent advantages in endoscopic tasks.
\end{itemize}

\begin{figure*}[ht]
	\centering 
    \vspace{-0.3cm}
	\includegraphics[width=\textwidth]{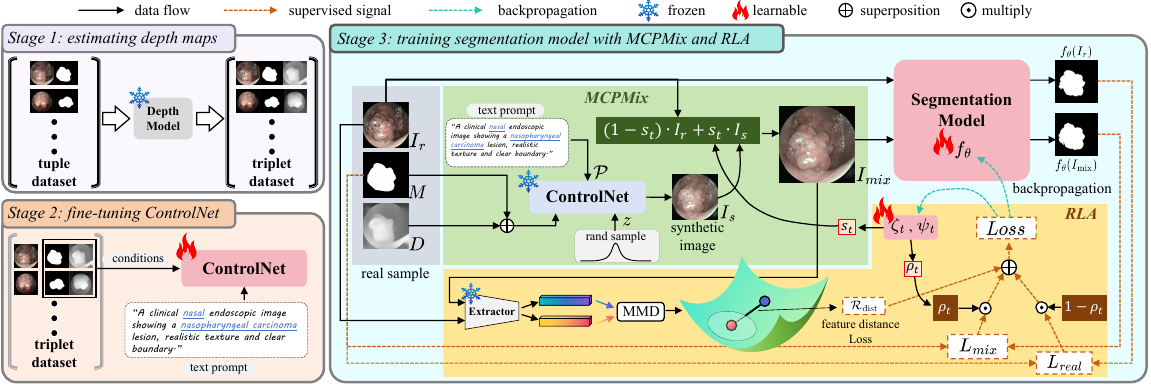}
    \vspace{-0.6cm}
	\caption{The proposed three-stage pipeline. Stage 1: a frozen depth network provides priors. Stage 2: diffusion-guided ControlNet is trained with mask, depth, and text, then frozen. Stage 3: MCPMix aims to preserve semantics and boundaries while expanding appearance diversity. RLA adaptively reduces both mixing strength and the loss weight of mixed samples, transitioning from strong exploration to real-domain.}
	\label{fig:overall_framework}
    \vspace{-0.5 cm} 
\end{figure*} 

\section{Related Work}\label{sec_relatedwork}

\subsection{Classical Imaging Augmentation}

Classical augmentation techniques mainly include geometric and photometric transformations applied to individual image–mask pairs, such as flipping, rotation, scaling, and intensity adjustment~\cite{AutoAugment_2019_CVPR}. These operations are simple, label-preserving, and widely used, but their diversity is inherently limited and often insufficient to model complex appearance variations in endoscopic scenes~\cite{classic_augmentation_shortness}.
Beyond single-sample transforms, inter-sample augmentation methods combine information from multiple images to enhance generalization~\cite{mixup_2018_ICLR,SmoothMix,PuzzleMix,CutMix,GridMix,HSMix}. Approaches inspired by sample interpolation or region substitution introduce new combinations of structures and textures. While such strategies can improve robustness in classification, they are less suited for dense prediction tasks, where inconsistent boundaries and soft labels may weaken pixel-level supervision~\cite{HOU2024110089,NEURIPS2022_e6f32e64}.
More recently, augmentation pipelines have been designed to enhance robustness to appearance shifts by composing diverse transformations or introducing texture-based perturbations~\cite{AugMix,PixMix}. Although these methods enrich style diversity and improve stability under distribution changes, they primarily operate in the image appearance domain and lack explicit semantic constraints, limiting their ability to preserve fine lesion boundaries critical for medical segmentation.
Compared with the above methods, we mix only in the appearance space, with hard labels and same-mask pairing, better preserving lesion geometry and boundary consistency.

\subsection{Generative Augmentation}

Diffusion-based augmentation with structural control, e.g., ControlNet, can generate images conditioned on masks for subsequent training, going beyond simple heuristic perturbations~\cite{ControlNet_ICCV2023}.
One line of work starts from a predicted or user-provided mask and generates an image conditioned on that mask~\cite{konz2024segguided}. 
Another category jointly synthesizes image-mask pairs~\cite{SatSynth}. 
A third category edits real images or inserts targets using a supplied mask~\cite{ControlPolypNet}. 
A fourth category changes style to broaden appearance coverage~\cite{DiffuseMix}.
Generative augmentation with diffusion can cut labeling cost and add variety~\cite{konz2024segguided,ControlPolypNet,DiffuseMix,SatSynth}. But two issues remain. First, current methods do not fully use mask-conditioned generation to make many appearances under the same mask. Second, synthetic images differ from real images in texture, color, and lighting, leading a distribution shift~\cite{Domain_Gap_2024_Nature,domain_gap_doge}. If training uses too many synthetic images, models may fit the synthetic style and generalize worse to real data~\cite{barbano2025steerable,Domain_Gap_2025_CVPR}.
In contrast, we integrate generative augmentation with inter-sample mixing to expand appearance diversity while preserving semantic geometry. To handle the synthetic-real shift, RLA places more weight to synthetic data early to learn diversity, then adaptively reduces this weight based on training backpropagation, guiding optimization toward the real-data distribution.

\section{Method}\label{sec_framework}
\subsection{Overview}

Our method is a diffusion-guided augmentation framework that comprises Mask-Consistent Paired Mixing (MCPMix) and Real-Anchored Learnable Annealing (RLA), which enriches appearance diversity under mask-consistent geometry with hard-label supervision and adaptively rebalances synthetic and real data to reduce domain shift and stabilize optimization.
Our method has three stages as shown in Fig.~\ref{fig:overall_framework}.
First, a frozen depth estimator computes depth maps for training images and provides structural priors that preserve the global layout of tissues and lesions.
Next, we fine tune ControlNet on real data with segmentation masks, cached depth, and short prompts to synthesize mask aligned images that increase appearance diversity while preserving lesion geometry.
Finally, we train the segmentation network with MCPMix and RLA.
Each real image is paired with its synthetic counterpart that shares the same mask, mixed at the input, and supervised by the original binary mask to avoid partial pixel ambiguity near boundaries.
RLA learns two scalars that control the input mixing ratio and the loss weight of synthetic samples.
A simple schedule with regularization uses more synthetic data early and then shifts back to real data to reduce overfitting to the synthetic domain.
The diffusion model is fine-tuned before segmentation training and kept frozen during segmentation.
Synthetic images are generated on-the-fly each epoch to provide new yet label consistent samples.
Notably, specular highlights and motion blur can still cause local label ambiguity and boundary uncertainty. Our pairing and hard label strategy reduces these effects but does not eliminate them.

\subsection{Preliminaries}
The training dataset consists of real endoscopic images and their pixel-wise masks, denoted by $\mathcal{D}=\{(I_r^{(n)},M^{(n)})\}_{n=1}^N$, where $I_r\colon\mathbb{R}^{H\times W\times C}$ is the real image and $M\in\{0,1\}^{H\times W}$ is the corresponding binary lesion mask.
A segmentation network $f_\theta\colon\mathbb{R}^{H\times W\times C}\to[0,1]^{H\times W}$ is trained to predict a pixel-wise probability map $\hat{M}$.
A synthetic image $I_s\colon\mathbb{R}^{H\times W\times C}$ is generated by a conditional generator $g_s$. 
Generation is conditioned on the real mask $M$ and structural priors such as a depth map $D$ and a short prompt $\mathcal{P}$, so that lesion geometry and location remain aligned while appearance varies, e.g., illumination, color, reflection, texture, device style, etc. Formally,
\begin{equation}
    \setlength{\abovedisplayskip}{3pt}
    \setlength{\belowdisplayskip}{3pt}
        \begin{aligned}
            I_s \sim g_s(M,D,\mathcal{P},z),   
    \end{aligned}
\end{equation}
where $z$ denotes the diffusion sampling noise, and the depth map $D$ for each image is estimated by a frozen pretrained depth estimator (DPT~\cite{Depth_Estimation}). 
The depth prior is used only in the synthesis pipeline and not during segmentation inference. 
$g_s$ is implemented as ControlNet~\cite{ControlNet_ICCV2023} initialized from Stable Diffusion~\cite{SDV15} and then fine-tuned.
This mask-consistent yet appearance-diverse design yields triplets $\mathcal{S}=\{(I_r,I_s,M)\}$, where each real sample may be paired with multiple synthetic counterparts to broaden appearance coverage.
On this basis, we innovatively introduce two complementary components: Mask-Consistent Paired Mixing (MCPMix) and Real-Anchored Learnable Annealing (RLA).

\subsection{Mask-Consistent Paired Mixing (MCPMix)}\label{subsec:MCPMix}
We construct mixed samples for any $(I_r,I_s,M)\in\mathcal{S}$ as
\begin{equation} \label{eq:mixup}
    \setlength{\abovedisplayskip}{3pt}
    \setlength{\belowdisplayskip}{3pt}
        \begin{aligned}
    I_{\mathrm{mix}} = (1-s_t)I_r + s_t I_s,
    \end{aligned}
\end{equation}
where $s_t \in [0,s_{\max}]$ is a learnable weight produced by RLA at step $t$ (see Sec.~\ref{subsec:RLA}). The upper bound $s_{\max}\le1$ caps the synthetic share. Unlike classical mixing, we do not soften the label by $s_t$. Instead, we use the ground-truth mask, $Y_{\mathrm{mix}}=M$, to supervise training.
Because $I_s$ and $I_r$ are aligned in lesion geometry and location, each pixel in $I_{\mathrm{mix}}$ is supervised by $M$. 
This reduces semantic ambiguity and label conflicts, keeping semantics stable.
Intuitively, MCPMix interpolates only in appearance, while the semantics are preserved.
During training, the model is supervised by both real and mixed samples. The segmentation loss $\ell_{\mathrm{seg}}(\hat{M},M)$ uses binary cross-entropy to promote stable pixel-wise convergence:
\begin{equation}
    \setlength{\abovedisplayskip}{3pt}
    \setlength{\belowdisplayskip}{4pt}
        \begin{aligned}
        L_{\mathrm{real}}&=\ell_{\mathrm{seg}}(f_\theta(I_r),M), 
    \end{aligned}
\end{equation}
\begin{equation}
    \setlength{\abovedisplayskip}{3pt}
    \setlength{\belowdisplayskip}{5pt}
        \begin{aligned}
        L_{\mathrm{mix}}&=\ell_{\mathrm{seg}}(f_\theta(I_{\mathrm{mix}}),M).  
    \end{aligned}
\end{equation}
MCPMix reduces the harmful effects of classical mixing by keeping hard labels. In classical mixing, two real images $I_a$ and $I_b$ are blended as $I'=\lambda I_a+(1-\lambda)I_b$, and the masks are mixed as $\lambda M_a+(1-\lambda)M_b$. At the pixel level, $M_a$ and $M_b$ often have misaligned boundaries or conflicting regions, so the supervision is no longer strictly binary. 
This leads to unstable gradients and encourages learning of blurred boundaries (see Sec.~\ref{subsec:why_gradient}). In contrast, MCPMix uses a shared mask $M$ for both the real and the synthetic image, thereby reducing soft label ambiguity caused by geometric mismatch.
MCPMix also avoids the structural issues in cut-and-paste mixing. Such methods can cut off or break up lesions or organs, harming anatomical integrity. By sharing the same mask and mixing whole images, MCPMix keeps structures intact and helps the model learn large, continuous targets.

\subsection{Real-Anchored Learnable Annealing (RLA)}\label{subsec:RLA}
RLA aims to adaptively balance the benefits of using synthetic samples with the need to keep the model close to the real-data distribution during training. To this end, we introduce two learnable, differentiable scalars. The first, $\rho_t$, controls the loss weight for mixed samples, and the second, $s_t$, controls the input mixing ratio. 
For stability and differentiability, we adopt a sigmoid parameterization:
\begin{equation}\label{eq:rho_t_s_t}
\setlength{\abovedisplayskip}{3pt}
\setlength{\belowdisplayskip}{3pt}
\begin{aligned}
\rho_t=\rho_{\max}\,\sigma(\psi_t),~ s_t=s_{\max}\,\sigma(\zeta_t), 
\end{aligned}
\end{equation}
where $\psi_t$ and $\zeta_t$ are learnable scalar outputs from a lightweight auxiliary network jointly trained with $\theta$. $\sigma(\cdot)$ denotes the sigmoid function and $\rho_{\max}, s_{\max}\in(0,1]$ are configurable upper bounds. 
This parameterization ensures $\rho_t\in[0,\rho_{\max}]$ and $s_t\in[0,s_{\max}]$ without extra projection or clipping, while remaining differentiable.

We use maximum mean discrepancy (MMD)~\cite{MMD} to measure the distributional discrepancy between real and mixed images. Let $\phi(\cdot)$ be a frozen feature extractor. Let $F_m=\phi(I_{\mathrm{mix}})$ and $F_r=\phi(I_r)$. The distributional discrepancy is $D_t=\mathrm{MMD}(F_m,F_r)$. Rather than minimizing $D_t$ directly, we use a soft margin: a penalty is added only when $D_t$ exceeds a dynamic threshold $\tau_t$,
\begin{equation}
    \setlength{\abovedisplayskip}{3pt}
    \setlength{\belowdisplayskip}{3pt}
    \begin{aligned}
    \mathcal{R}_{\mathrm{dist}}=\mu  [D_t-\tau_t]_+,  
    \end{aligned}
\end{equation}
where $[x]_+=\max(x,0)$. The threshold $\tau_t$ is gradually tightened during training, which guides the distribution of the mixed samples toward the real domain in later stages. We define $\tau_t$ with a cosine annealing schedule:
\begin{equation}
    \setlength{\abovedisplayskip}{3pt}
    \setlength{\belowdisplayskip}{3pt}
    \begin{aligned}
    \tau_t=\tau_0 \frac{1+\cos(\pi t/T)}{2},  
    \end{aligned}
\end{equation}
with $T$ denoting the total number of epochs. Intuitively, this allows larger early-stage discrepancies to encourage synthetic diversity, then progressively reduces tolerance to align mixed samples with the real distribution, thereby mitigating synthetic-domain bias. The distribution constraint propagates gradients through MMD to the input $I_{mix}$ and, via Eq.~(\ref{eq:mixup}) and Eq.~(\ref{eq:rho_t_s_t}), further back to $\zeta_t$.

For the supervision loss, $\rho_t$ regulates the trade-off between real and mixed samples, while both $\rho_t$ and $s_t$ are further regularized by mild temporal priors, denoted $\rho_t^{\mathrm{prior}}$ and $s_t^{\mathrm{prior}}$. Specifically, these priors follow cosine-annealing schedules, serving as weak trend guidance and regularization to prevent oscillations or collapse during the early and middle stages of training. The overall objective is expressed as
\begin{equation}
    \setlength{\abovedisplayskip}{3pt}
    \setlength{\belowdisplayskip}{3pt}
    \begin{aligned}
            L_t(\theta,\psi_t,\zeta_t)
            &=(1-\rho_t)L_{\mathrm{real}}+\rho_t L_{\mathrm{mix}}
            +\mu[D_t-\tau_t]_+ \\
            &\quad+\lambda_\rho(\rho_t-\rho_t^{\mathrm{prior}})^2
            +\lambda_s(s_t-s_t^{\mathrm{prior}})^2,
    \end{aligned}
\end{equation}
where $\lambda_\rho,\lambda_s>0$ are the prior regularization coefficients, set to $10^{-3}$. These priors do not impose strong constraints on $\rho_t$ and $s_t$. Instead, they provide weak guidance, ensuring that the dominant driving force arises from data-driven gradients rather than predefined schedules.
Algorithm~\ref{alg:A} summarizes the training procedure of MCPMix and RLA.
\begin{figure}[!t]
     \vspace{-0.3cm}
    \centering
    \removelatexerror
    \begin{algorithm}[H]
        \caption{\fontsize{8.5}{18}\selectfont Our training schedule (MCPMix+RLA).}
        \label{alg:A}
        \footnotesize 
        \begin{algorithmic}[1]
            \REQUIRE Dataset $\mathcal{D}=\{(I_r,M)\}$; synthesizer $g_s$; frozen encoder $\phi$; segmentation model $f_\theta$; priors $(D,p)$; hyper-params $\rho_{\max}, s_{\max}, \mu, \lambda_\rho, \lambda_s, \tau_0$; batch size $B$; epochs $T$.
            \ENSURE Trained $f_\theta$; 
            \STATE Initialize $\theta$; set gates $\psi \gets 0$, $\zeta \gets 0$
            \FOR{$t=1$ \textbf{to} $T+300$}
                \STATE $\tau_t \gets \tau_0 \cdot \frac{1+\cos(\pi t/T)}{2}$;\quad
                       $\rho_t \gets \rho_{\max}\sigma(\psi)$;\quad
                       $s_t \gets s_{\max}\sigma(\zeta)$
                \FOR{each minibatch $\{(I_r^{(i)},M^{(i)})\}_{i=1}^{B}$}
                    \STATE Generate $I_s^{(i)} \sim g_s(M^{(i)};D,\mathcal{P},z)$
                    \STATE $I_{\mathrm{mix}} \gets (1-s_t)\, I_r + s_t\, I_s$ 
                    \STATE $\hat{M}_r \gets f_\theta(I_r)$;\quad $\hat{M}_m \gets f_\theta(I_{\mathrm{mix}})$
                    \STATE $L_{\mathrm{real}} \gets \ell_{\mathrm{seg}}(\hat{M}_r,M)$;\quad
                           $L_{\mathrm{mix}} \gets \ell_{\mathrm{seg}}(\hat{M}_m,M)$
                    \STATE $F_r \gets \phi(I_r)$;\quad $F_m \gets \phi(I_{\mathrm{mix}})$ 
                    \STATE $D_t \gets \mathrm{MMD}(F_m,F_r)$
                    \STATE $\mathcal{R}_{\mathrm{dist}} \gets \mu \cdot \max(0, D_t - \tau_t)$;\quad
                           $\text{prior} \gets \lambda_\rho(\rho_t-\rho_t^{\mathrm{prior}})^2 + \lambda_s(s_t-s_t^{\mathrm{prior}})^2$
                    \STATE $L \gets (1-\rho_t)L_{\mathrm{real}} + \rho_t L_{\mathrm{mix}} + \mathcal{R}_{\mathrm{dist}} + \text{prior}$
                    \STATE Update $\theta,\psi,\zeta$ by backpropagation; 
                \ENDFOR  \COMMENT{stop after $\lceil|\mathcal{D}|/B\rceil$ batches}

            \ENDFOR
        \end{algorithmic}
    \end{algorithm}
  \vspace{-1.00 cm}
\end{figure}

\subsection{Differentiability and Gradient Flow of RLA}
We further provide a differentiability analysis to demonstrate that the parameters $\psi_t$ and $\zeta_t$ can indeed be updated through backpropagation.
For $\psi_t$, since $D_t$ and $s_t$ do not directly depend on $\rho_t$ (with $D_t$ depending on $I_{\text{mix}}$, which itself depends only on $s_t$), we obtain
\begin{equation}\nonumber
    \setlength{\abovedisplayskip}{3pt}
    \setlength{\belowdisplayskip}{3pt}
    \begin{aligned}
\frac{\partial \mathcal{L}_t}{\partial \rho_t}
= -L_{\text{real}} + L_{\text{mix}} + 2\lambda_\rho(\rho_t - \rho_t^{\text{prior}}).
    \end{aligned}
\end{equation}
Given $\rho_t = \rho_{\max}\sigma(\psi_t)$, it follows that
\begin{equation}\nonumber
    \setlength{\abovedisplayskip}{3pt}
    \setlength{\belowdisplayskip}{3pt}
    \begin{aligned}
\frac{\partial \mathcal{L}_t}{\partial \psi_t}
=\Big(-L_{\text{real}} + L_{\text{mix}} + 2\lambda_\rho(\rho_t - \rho_t^{\text{prior}})\Big)\cdot \rho_{\max}~\sigma'(\psi_t).
    \end{aligned}
\end{equation}
For $\zeta_t$, the analysis is as follows. The real-sample loss $L_{\text{real}}$ is independent of $s_t$, thus its derivative vanishes. The gradient of the mixed-sample loss is
\begin{equation}\nonumber
    \setlength{\abovedisplayskip}{3pt}
    \setlength{\belowdisplayskip}{3pt}
    \begin{aligned}
\frac{\partial L_{\text{mix}}}{\partial s_t}
=\Big\langle \frac{\partial L_{\text{mix}}}{\partial I_{\text{mix}}},~ I_s-I_r \Big\rangle.
    \end{aligned}
\end{equation}
When $D_t > \tau_t$, the distribution discrepancy term contributes
\begin{equation}\nonumber
    \setlength{\abovedisplayskip}{3pt}
    \setlength{\belowdisplayskip}{3pt}
    \begin{aligned}
\frac{\partial D_t}{\partial s_t}
=\Big\langle \frac{\partial D_t}{\partial I_{\text{mix}}},~ I_s-I_r \Big\rangle.
    \end{aligned}
\end{equation}
Combining all terms, the derivative with respect to $s_t$ is
\begin{equation}\nonumber
    \setlength{\abovedisplayskip}{3pt}
    \setlength{\belowdisplayskip}{3pt}
    \begin{aligned}
\frac{\partial \mathcal{L}_t}{\partial s_t}
&=\rho_t \Big\langle \frac{\partial L_{\text{mix}}}{\partial I_{\text{mix}}}, I_s-I_r\Big\rangle \\
+~\mu~\mathbf{1}_{\{D_t>\tau_t\}} &\Big\langle \frac{\partial D_t}{\partial I_{\text{mix}}}, I_s-I_r\Big\rangle 
+2\lambda_s(s_t-s_t^{\text{prior}}).
    \end{aligned}
\end{equation}
Since $s_t = s_{\max}\sigma(\zeta_t)$, we further have
\begin{equation}\nonumber
    \setlength{\abovedisplayskip}{3pt}
    \setlength{\belowdisplayskip}{3pt}
    \begin{aligned}
\frac{\partial \mathcal{L}_t}{\partial \zeta_t}
=\frac{\partial \mathcal{L}_t}{\partial s_t}\cdot s_{\max}~\sigma'(\zeta_t).
    \end{aligned}
\end{equation}
In conclusion, both $\psi_t$ and $\zeta_t$ are fully differentiable and can be optimized via standard backpropagation, thereby ensuring their learnability during training.

\section{Experiment}\label{sec_experiment}

\subsection{Datasets}\label{subsec_dataset}
\paragraph{Public datasets}
We benchmark on community datasets for comparability and reproducibility.
For gastrointestinal endoscopy, we use Kvasir\text-SEG~\cite{Kvasir_SEG} with 1{,}000 polyp images, PICCOLO~\cite{PICCOLO} with 3{,}433 multi-center images, and CVC\text-ClinicDB~\cite{CVC_ClinicDB_1} with 612 images.
To assess transfer beyond endoscopy, we report results on ISIC 2017~\cite{ISIC2017} with 2{,}600 dermoscopic images and binary expert lesion masks.
All datasets follow community split protocols.

\paragraph{Private clinical dataset}
We further evaluate on NPC\text-LES 2023~\cite{jiepy_TCSVT}, a nasal endoscopic dataset for nasopharyngeal carcinoma segmentation collected at the First Affiliated Hospital of Sun Yat\text-sen University with ethics approval and written informed consent.
The dataset contains 3{,}182 training images and 453 test images with pixel-level labels.
Splits are patient-level to prevent leakage from adjacent frames.

\subsection{Implementation Details}
\textbf{Stage 1.}
We estimate a depth map for each training image using DPT. Depth serves only as a conditioning signal in the next stages and receives no gradients.
\textbf{Stage 2.}
We fine tune ControlNet on Stable Diffusion v1.5 with the ground truth mask, the cached depth, and a short text prompt to synthesize mask aligned images. 
The prompt is:
\emph{A clinical \{IMAGING POSITION\} endoscopic image showing a \{TARGET\} lesion, realistic texture and clear boundary.}
\textbf{Stage 3.}
We train SegFormer~\cite{SegFormer} in PyTorch with $512{\times}512$ inputs using AdamW with learning rate $1\times10^{-3}$ and weight decay $1\times10^{-4}$. The total budget is $T{=}400$ epochs with effective batch size $32$. ControlNet remains frozen, and synthetic images are generated on-the-fly by data loader workers. Each real image is paired with its ControlNet counterpart conditioned on the same mask, and we apply full image linear mixing with supervision from the original mask. In RLA, we learn $s_t$ and $\rho_t$ with upper bounds $s_{\max}{=}0.7$ and $\rho_{\max}{=}0.5$. MMD is computed in a frozen ResNet 50 feature space with a Gaussian RBF kernel.

\subsection{Evaluation metrics}
We report five standard metrics consistent with~\cite{jiepy_TCSVT,xiang2024lightweight}: mean Intersection over Union ($\text{mIoU}$), $\text{Precision}$, $\text{Recall}$, Pixel Accuracy ($\text{PA}$), and Dice Similarity Coefficient ($\text{DSC}$). Here $TP,FP,FN,TN$ are pixel counts and $k$ indicates two classes i.e., foreground and background. Formally:
\begin{equation}
    \setlength{\abovedisplayskip}{3pt}
    \setlength{\belowdisplayskip}{3pt}
    \resizebox{0.68\hsize}{!}{$\begin{aligned}
    \mathrm{mIoU}&=\frac{1}{k}\sum_{i=1}^{k}\frac{TP_i}{TP_i+FP_i+FN_i}\times 100\% ,\\
    \mathrm{Precision}&=\frac{TP}{TP+FP}\times 100\% ,\\
    \mathrm{Recall}&=\frac{TP}{TP+FN}\times 100\% ,\\
    \mathrm{PA}&=\frac{TP+TN}{TP+TN+FP+FN}\times 100\% ,\\
    \mathrm{DSC}&=\frac{2TP}{2TP+FP+FN}\times 100\% .
    \end{aligned}$}
\end{equation}
We report $\text{HD}_{95}$, ASSD~\cite{HD95_ASSD}, boundary Precision, Recall, and F1 ($\text{B-P}$, $\text{B-R}$, $\text{B-F1}$)~\cite{BF1score}, and Boundary IoU (BIoU)~\cite{BIOU} at tolerances $\delta\in\{2,5,10\}$ pixels. Let $y,\hat{y}\in\{0,1\}^{H\times W}$ denote the ground truth and the prediction, $\partial y,\partial\hat{y}$ their pixel boundaries, and $d(p,\partial y)$ the Euclidean distance in pixels from a pixel $p$ to $\partial y$. Then
\begin{equation}
    \setlength{\abovedisplayskip}{3pt}
    \setlength{\belowdisplayskip}{3pt}
    \resizebox{0.88\hsize}{!}{$\begin{aligned}
    \text{HD}_{95}&=\operatorname{perc}_{95}\!\Big(\{\,d(p,\partial \hat{y})\mid p\!\in\!\partial y\,\}\cup\{\,d(p,\partial y)\mid p\!\in\!\partial \hat{y}\,\}\Big),\\
    \text{ASSD}&=\tfrac12\!\Big(\frac{1}{|\partial y|}\!\sum_{p\in\partial y}\! d(p,\partial \hat{y})+\frac{1}{|\partial \hat{y}|}\!\sum_{g\in\partial \hat{y}}\! d(g,\partial y)\Big),\\
    \mathrm{B\textrm{-}R}&=\frac{1}{|\partial y|}\sum_{p\in\partial y} 1 [\,d(p,\partial \hat{y})\le \delta\,],\\
    \mathrm{B\textrm{-}P}&=\frac{1}{|\partial \hat{y}|}\sum_{p\in\partial \hat{y}} 1 [\,d(p,\partial y)\le \delta\,],~
    \mathrm{B\textrm{-}F1} =\frac{2\,\mathrm{B\textrm{-}P}\,\mathrm{B\textrm{-}R}}{\mathrm{B\textrm{-}P}+\mathrm{B\textrm{-}R}}.
    \end{aligned}$}
\end{equation}
For BIoU, define the $r$-neighborhood (boundary band) of a boundary as
$\mathcal{B}_y^{(r)}=\{\,p\in\Omega:\,d(p,\partial y)\le r\,\}$ and
$\mathcal{B}_{\hat{y}}^{(r)}=\{\,p\in\Omega:\,d(p,\partial \hat{y})\le r\,\}$. The boundary IoU is
\begin{equation}
    \setlength{\abovedisplayskip}{3pt}
    \setlength{\belowdisplayskip}{3pt}
    \resizebox{0.45\hsize}{!}{$\begin{aligned}
    \mathrm{BIoU}=\frac{\left|\mathcal{B}_y^{(r)}\cap \mathcal{B}_{\hat{y}}^{(r)}\right|}{\left|\mathcal{B}_y^{(r)}\cup \mathcal{B}_{\hat{y}}^{(r)}\right|}\in[0,1].
    \end{aligned}$}
\end{equation}
Our method is expected to lower $\text{HD}_{95}$ and ASSD while increasing $\text{B-F1}$ and $\mathrm{BIoU}$ ($r=2$px).

\subsection{Comparison with SOTA}
\subsubsection{SOTA Methods}
We compare our method with four families of mixing-based baselines: (i) global linear mixing (Mixup~\cite{mixup_2018_ICLR}, SmoothMix~\cite{SmoothMix}); (ii) copy-and-paste (CutMix~\cite{CutMix}, GridMix~\cite{GridMix}, PuzzleMix~\cite{PuzzleMix}, HSMix~\cite{HSMix}); (iii) multi-augmentation blending for robustness (AugMix~\cite{AugMix}, PixMix~\cite{PixMix}); and (iv) real–synthetic composition with generative appearance perturbation (DiffuseMix~\cite{DiffuseMix}). These cover classical mixing methods, forming complementary and competitive baselines. For fairness, we adopt open source implementations when available and otherwise reimplement them under a unified recipe with search ranges following prior recommendations.

We also compare our method with diffusion driven data augmentation that synthesizes training samples (Sec.~\ref{subsec:Generative}).
The suite includes mask conditioned editing on real images with ControlPolypNet~\cite{ControlPolypNet}, mask first synthesis with GenSRRFI~\cite{GenSRRFI}, and joint image mask generation with SatSynth~\cite{SatSynth}.
All methods are retrained under a unified setting with the same backbone and input resolution to ensure fair and reproducible comparison.

\begin{table}[t]
    \belowrulesep=0pt
    \aboverulesep=0pt
    \caption{\sm{7}Comparison with mixing-based methods on Kvasir-SEG, PICCOLO, CVC ClinicDB and NPC-LES (Mean \text{\sm{6}$\pm$Standard deviation}, n=4).}
    \label{Table:different_framework__Kvasir_PICCOLO_CVC_NPC}
    \centering
    \setlength{\tabcolsep}{0pt}
    \begin{tabular*}{\linewidth}{@{}@{\extracolsep{\fill}}c|cccccc}
        \toprule
        \sm{7}Dataset&
        \sm{7}Methods&
        \sm{7}$\text{mIoU}$ $ \left( \% \right) $&
        \sm{7}$\text{PA}$ $ \left( \% \right) $ &
        \sm{7}$\text{Recall}$ $ \left( \% \right) $ &
        \sm{7}$\text{Precision}$ $ \left( \% \right) $ &
        \sm{7}$\text{DSC}$$ \left( \% \right) $  \\
        \hline
        \multirow{11}{*}{\makecell{\sm{7} Kvasir \\ \sm{7} -SEG}}
        &\sm{7}CutMix\sm{6}\cite{CutMix} 
            &\sm{7}86.76\sm{5}$\pm$0.95&\sm{7}96.02\sm{5}$\pm$0.62&\sm{7}85.85\sm{5}$\pm$1.14&\sm{7}91.13\sm{5}$\pm$0.39&\sm{7}86.24\sm{5}$\pm$0.91\\
        &\sm{7}Mixup\sm{6}\cite{mixup_2018_ICLR} 
            &\sm{7}85.16\sm{5}$\pm$0.98&\sm{7}95.36\sm{5}$\pm$0.93&\sm{7}83.78\sm{5}$\pm$1.01&\sm{7}90.37\sm{5}$\pm$0.73&\sm{7}84.59\sm{5}$\pm$0.64\\
        &\sm{7}GridMix\sm{6}\cite{GridMix}  
            &\sm{7}80.25\sm{5}$\pm$1.09&\sm{7}93.90\sm{5}$\pm$1.18&\sm{7}72.92\sm{5}$\pm$1.56&\sm{7}{92.63}\sm{5}$\pm$1.24&\sm{7}77.35\sm{5}$\pm$0.91\\
        &\sm{7}SmoothMix\sm{6}\cite{SmoothMix}  
            &\sm{7}86.82\sm{5}$\pm$1.70&\sm{7}96.12\sm{5}$\pm$0.72&\sm{7}86.30\sm{5}$\pm$0.52&\sm{7}90.87\sm{5}$\pm$0.83&\sm{7}85.97\sm{5}$\pm$1.31\\
        &\sm{7}PuzzleMix\sm{6}\cite{PuzzleMix} 
            &\sm{7}86.80\sm{5}$\pm$1.55&\sm{7}95.94\sm{5}$\pm$1.18&\sm{7}86.39\sm{5}$\pm$1.32&\sm{7}90.47\sm{5}$\pm$0.48&\sm{7}86.61\sm{5}$\pm$1.23\\
        &\sm{7}AugMix\sm{6}\cite{AugMix} 
            &\sm{7}86.68\sm{5}$\pm$0.81&\sm{7}95.88\sm{5}$\pm$0.80&\sm{7}85.29\sm{5}$\pm$0.53&\sm{7}91.01\sm{5}$\pm$1.19&\sm{7}86.24\sm{5}$\pm$0.33\\
        &\sm{7}PixMix\sm{6}\cite{PixMix}  
            &\sm{7}86.72\sm{5}$\pm$1.63&\sm{7}95.80\sm{5}$\pm$0.81&\sm{7}84.49\sm{5}$\pm$0.65&\sm{7}92.17\sm{5}$\pm$0.85&\sm{7}86.78\sm{5}$\pm$1.27\\
        &\sm{7}DiffuseMix\sm{6}\cite{DiffuseMix}  
            &\sm{7}{87.60}\sm{5}$\pm$1.02&\sm{7}{96.24}\sm{5}$\pm$0.71&\sm{7}{86.55}\sm{5}$\pm$1.08&\sm{7}91.35\sm{5}$\pm$0.75&\sm{7}{87.43}\sm{5}$\pm$0.96\\
        &\sm{7}HSMix\sm{6}\cite{HSMix}  
            &\sm{7}86.77\sm{5}$\pm$1.54&\sm{7}96.00\sm{5}$\pm$0.67&\sm{7}86.18\sm{5}$\pm$1.45&\sm{7}90.50\sm{5}$\pm$0.82&\sm{7}86.02\sm{5}$\pm$1.05\\
        &\sm{7}\bf{Ours}
            &\sm{7}\best{88.72}\sm{5}$\pm$0.30&\sm{7}\best{96.55}\sm{5}$\pm$0.75&\sm{7}\best{87.18}\sm{5}$\pm$0.66&\sm{7}\best{93.21}\sm{5}$\pm$0.77&\sm{7}\best{88.13}\sm{5}$\pm$0.33\\
        \hline
        \multirow{11}{*}{\makecell{\sm{7} PICCOLO}}
        &\sm{7}CutMix\sm{6}\cite{CutMix} 
            &\sm{7}81.52\sm{5}$\pm$1.03&\sm{7}95.61\sm{5}$\pm$0.59&\sm{7}73.70\sm{5}$\pm$1.02&\sm{7}90.48\sm{5}$\pm$0.91&\sm{7}77.99\sm{5}$\pm$1.38\\
        &\sm{7}Mixup\sm{6}\cite{mixup_2018_ICLR}  
            &\sm{7}80.19\sm{5}$\pm$1.01&\sm{7}94.29\sm{5}$\pm$0.70&\sm{7}71.82\sm{5}$\pm$0.96&\sm{7}91.13\sm{5}$\pm$0.64&\sm{7}76.46\sm{5}$\pm$1.04\\
        &\sm{7}GridMix\sm{6}\cite{GridMix}  
            &\sm{7}68.99\sm{5}$\pm$0.75&\sm{7}90.66\sm{5}$\pm$0.81&\sm{7}51.02\sm{5}$\pm$0.53&\sm{7}87.57\sm{5}$\pm$0.71&\sm{7}63.43\sm{5}$\pm$0.62\\
        &\sm{7}SmoothMix\sm{6}\cite{SmoothMix}  
            &\sm{7}82.03\sm{5}$\pm$1.17&\sm{7}{95.96}\sm{5}$\pm$0.66&\sm{7}73.01\sm{5}$\pm$0.22&\sm{7}90.10\sm{5}$\pm$1.12&\sm{7}79.16\sm{5}$\pm$0.19\\
        &\sm{7}PuzzleMix\sm{6}\cite{PuzzleMix}  
            &\sm{7}79.08\sm{5}$\pm$0.73&\sm{7}94.86\sm{5}$\pm$0.73&\sm{7}67.84\sm{5}$\pm$1.07&\sm{7}\best{94.15}\sm{5}$\pm$0.51&\sm{7}72.30\sm{5}$\pm$0.99\\
        &\sm{7}AugMix\sm{6}\cite{AugMix}  
            &\sm{7}{83.08}\sm{5}$\pm$0.44&\sm{7}95.31\sm{5}$\pm$0.60&\sm{7}\best{82.03}\sm{5}$\pm$1.49&\sm{7}84.87\sm{5}$\pm$1.54&\sm{7}{81.84}\sm{5}$\pm$0.95\\
        &\sm{7}PixMix\sm{6}\cite{PixMix}  
            &\sm{7}81.53\sm{5}$\pm$0.78&\sm{7}94.91\sm{5}$\pm$0.73&\sm{7}76.58\sm{5}$\pm$0.27&\sm{7}85.69\sm{5}$\pm$0.74&\sm{7}79.18\sm{5}$\pm$0.62\\
        &\sm{7}DiffuseMix\sm{6}\cite{DiffuseMix}  
            &\sm{7}78.67\sm{5}$\pm$0.41&\sm{7}93.08\sm{5}$\pm$0.67&\sm{7}70.09\sm{5}$\pm$0.57&\sm{7}90.18\sm{5}$\pm$0.32&\sm{7}75.10\sm{5}$\pm$0.98\\
        &\sm{7}HSMix\sm{6}\cite{HSMix}  
            &\sm{7}81.27\sm{5}$\pm$0.70&\sm{7}95.74\sm{5}$\pm$0.66&\sm{7}70.91\sm{5}$\pm$0.24&\sm{7}91.75\sm{5}$\pm$1.68&\sm{7}77.01\sm{5}$\pm$0.94\\
        &\sm{7}\bf{Ours}
            &\sm{7}\best{87.11}\sm{5}$\pm$0.59&\sm{7}\best{97.62}\sm{5}$\pm$0.34&\sm{7}{81.27}\sm{5}$\pm$0.78&\sm{7}{92.94}\sm{5}$\pm$0.42&\sm{7}\best{84.24}\sm{5}$\pm$0.49\\
        \hline
        \multirow{11}{*}{\makecell{\sm{7}CVC\\ \sm{7}ClinicDB}}
        &\sm{7}CutMix\sm{6}\cite{CutMix} 
            &\sm{7}90.87\sm{5}$\pm$2.40&\sm{7}98.62\sm{5}$\pm$0.25&\sm{7}87.32\sm{5}$\pm$0.77&\sm{7}93.38\sm{5}$\pm$0.91&\sm{7}91.15\sm{5}$\pm$0.80\\
        &\sm{7}Mixup\sm{6}\cite{mixup_2018_ICLR}  
            &\sm{7}86.33\sm{5}$\pm$1.00&\sm{7}98.09\sm{5}$\pm$0.29&\sm{7}77.75\sm{5}$\pm$0.90&\sm{7}\best{95.55}\sm{5}$\pm$1.32&\sm{7}83.59\sm{5}$\pm$0.56\\
        &\sm{7}GridMix\sm{6}\cite{GridMix}  
            &\sm{7}79.30\sm{5}$\pm$1.22&\sm{7}96.87\sm{5}$\pm$0.45&\sm{7}66.95\sm{5}$\pm$0.58&\sm{7}86.49\sm{5}$\pm$1.33&\sm{7}75.92\sm{5}$\pm$0.64\\
        &\sm{7}SmoothMix\sm{6}\cite{SmoothMix}  
            &\sm{7}{91.43}\sm{5}$\pm$1.16&\sm{7}{98.69}\sm{5}$\pm$0.18&\sm{7}{88.33}\sm{5}$\pm$1.23&\sm{7}94.37\sm{5}$\pm$0.56&\sm{7}91.23\sm{5}$\pm$1.12\\
        &\sm{7}PuzzleMix\sm{6}\cite{PuzzleMix}  
            &\sm{7}82.39\sm{5}$\pm$1.12&\sm{7}97.04\sm{5}$\pm$0.64&\sm{7}71.65\sm{5}$\pm$0.55&\sm{7}93.74\sm{5}$\pm$0.96&\sm{7}78.93\sm{5}$\pm$0.12\\
        &\sm{7}AugMix\sm{6}\cite{AugMix}  
            &\sm{7}90.88\sm{5}$\pm$0.99&\sm{7}98.62\sm{5}$\pm$0.24&\sm{7}87.10\sm{5}$\pm$1.25&\sm{7}93.46\sm{5}$\pm$1.22&\sm{7}{91.51}\sm{5}$\pm$1.01\\
        &\sm{7}PixMix\sm{6}\cite{PixMix}  
            &\sm{7}88.92\sm{5}$\pm$0.48&\sm{7}98.38\sm{5}$\pm$0.24&\sm{7}85.13\sm{5}$\pm$0.94&\sm{7}91.28\sm{5}$\pm$1.22&\sm{7}88.91\sm{5}$\pm$0.51\\
        &\sm{7}DiffuseMix\sm{6}\cite{DiffuseMix}  
            &\sm{7}89.13\sm{5}$\pm$0.55&\sm{7}98.25\sm{5}$\pm$0.42&\sm{7}84.01\sm{5}$\pm$1.16&\sm{7}93.54\sm{5}$\pm$0.66&\sm{7}88.60\sm{5}$\pm$0.60\\
        &\sm{7}HSMix\sm{6}\cite{HSMix}  
            &\sm{7}88.20\sm{5}$\pm$0.25&\sm{7}98.27\sm{5}$\pm$0.28&\sm{7}83.17\sm{5}$\pm$1.05&\sm{7}91.95\sm{5}$\pm$0.80&\sm{7}88.12\sm{5}$\pm$0.45\\
        &\sm{7}\bf{Ours}
            &\sm{7}\best{92.63}\sm{5}$\pm$0.36&\sm{7}\best{98.95}\sm{5}$\pm$0.05&\sm{7}\best{90.16}\sm{5}$\pm$0.92&\sm{7}{95.32}\sm{5}$\pm$0.71&\sm{7}\best{91.94}\sm{5}$\pm$0.60\\
        \hline
        \multirow{10}{*}{\makecell{\sm{7} NPC\\\sm{7} -LES }}
        &\sm{7}CutMix\sm{6}\cite{CutMix} 
            &\sm{7}86.47\sm{5}$\pm$0.76&\sm{7}94.21\sm{5}$\pm$0.25&\sm{7}88.94\sm{5}$\pm$1.36&\sm{7}92.18\sm{5}$\pm$1.93&\sm{7}88.33\sm{5}$\pm$2.14\\
        &\sm{7}Mixup\sm{6}\cite{mixup_2018_ICLR} 
            &\sm{7}85.97\sm{5}$\pm$1.20&\sm{7}94.05\sm{5}$\pm$0.52&\sm{7}89.11\sm{5}$\pm$1.58&\sm{7}91.52\sm{5}$\pm$1.74&\sm{7}87.30\sm{5}$\pm$1.80\\
        &\sm{7}GridMix\sm{6}\cite{GridMix} 
            &\sm{7}84.97\sm{5}$\pm$0.74&\sm{7}93.51\sm{5}$\pm$0.41&\sm{7}87.29\sm{5}$\pm$0.98&\sm{7}92.03\sm{5}$\pm$0.74&\sm{7}86.90\sm{5}$\pm$1.61\\
        &\sm{7}SmoothMix\sm{6}\cite{SmoothMix} 
            &\sm{7}86.79\sm{5}$\pm$0.15&\sm{7}94.44\sm{5}$\pm$0.14&\sm{7}{90.56}\sm{5}$\pm$0.38&\sm{7}90.86\sm{5}$\pm$0.51&\sm{7}88.64\sm{5}$\pm$0.23\\
        &\sm{7}PuzzleMix\sm{6}\cite{PuzzleMix} 
            &\sm{7}86.66\sm{5}$\pm$1.17&\sm{7}94.08\sm{5}$\pm$0.51&\sm{7}89.19\sm{5}$\pm$1.25&\sm{7}92.81\sm{5}$\pm$1.48&\sm{7}88.90\sm{5}$\pm$1.64\\
        &\sm{7}AugMix\sm{6}\cite{AugMix} 
            &\sm{7}{88.47}\sm{5}$\pm$0.58&\sm{7}{94.64}\sm{5}$\pm$0.34&\sm{7}90.06\sm{5}$\pm$1.19&\sm{7}\best{94.98}\sm{5}$\pm$0.85&\sm{7}{91.29}\sm{5}$\pm$0.95\\
        &\sm{7}PixMix\sm{6}\cite{PixMix} 
            &\sm{7}87.57\sm{5}$\pm$1.00&\sm{7}94.40\sm{5}$\pm$0.60&\sm{7}90.19\sm{5}$\pm$1.81&\sm{7}93.35\sm{5}$\pm$1.53&\sm{7}89.83\sm{5}$\pm$1.21\\
        &\sm{7}DiffuseMix\sm{6}\cite{DiffuseMix} 
            &\sm{7}88.01\sm{5}$\pm$0.79&\sm{7}94.52\sm{5}$\pm$0.43&\sm{7}90.41\sm{5}$\pm$1.28&\sm{7}93.92\sm{5}$\pm$1.12&\sm{7}90.46\sm{5}$\pm$0.96\\
        &\sm{7}HSMix\sm{6}\cite{HSMix} 
            &\sm{7}87.40\sm{5}$\pm$0.86&\sm{7}94.47\sm{5}$\pm$0.47&\sm{7}89.64\sm{5}$\pm$1.39&\sm{7}93.17\sm{5}$\pm$1.37&\sm{7}89.52\sm{5}$\pm$1.28\\
        &\sm{7}\bf{Ours}
            &\sm{7}\best{90.10}\sm{5}$\pm$0.66&\sm{7}\best{95.49}\sm{5}$\pm$0.41&\sm{7}\best{92.78}\sm{5}$\pm$1.20&\sm{7}{94.49}\sm{5}$\pm$0.83&\sm{7}\best{92.57}\sm{5}$\pm$0.37\\
        \bottomrule
    \end{tabular*}
    \vspace{-0.6cm}
\end{table}

\begin{figure}[!h]
    \centering
    \vspace{-0.05cm}
    \includegraphics[width=8.5 CM]{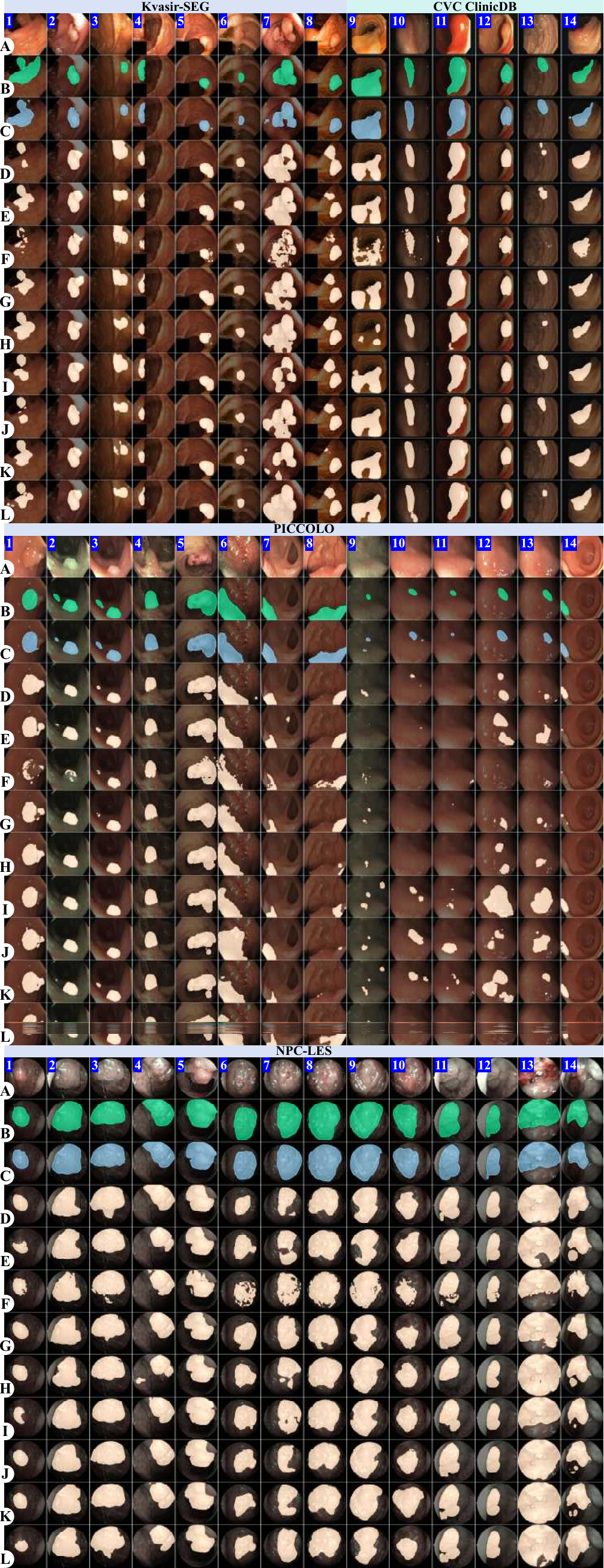}
    \caption{Visualization of different methods on Kvasir-SEG, CVC ClinicDB, PICCOLO and NPC-LES. (A:Image, B:\textbf{\textcolor[RGB]{0,235,140}{Ground-Truth}}, C:\textbf{\textcolor[RGB]{120,196,240}{Ours}}, D:CutMix, E:Mixup, F:GridMix, G:SmoothMix, H:PuzzleMix, I:Augmix, J:PixMix, K:DiffuseMix, L:HSMix)}
    \label{fig:Different_methods_Kvasir_CVC}
    \vspace{-0.4cm}
\end{figure} 
\subsubsection{Quantitative analysis}
Table~\ref{Table:different_framework__Kvasir_PICCOLO_CVC_NPC} summarizes all datasets in a single view. 
Across Kvasir-SEG, PICCOLO, CVC-ClinicDB, and NPC-LES, our method achieves the best overall segmentation quality with consistent gains in mIoU and DSC and competitive Precision–Recall balance. 
On Kvasir-SEG, our mIoU reaches 88.72\% with DSC 88.13\%, exceeding the strongest baseline by about one point while keeping low variance across runs. 
On PICCOLO, which is larger and more diverse, our gains are more pronounced, with mIoU 87.11\% and DSC 84.24\%, clearly ahead of mixing-based baselines. 
On CVC-ClinicDB we obtain the highest mIoU 92.63\% and DSC 91.94\%, with the best PA and the strongest Recall among competitors. 
On the private clinical cohort NPC-LES our method delivers mIoU 90.10\% and DSC 92.57\%, the highest among all methods, with the top Recall 92.78\% and near-top Precision 94.49\%. 
These trends match the design goal: mask-consistent appearance mixing preserves lesion geometry and boundary supervision while still widening appearance coverage. 
Global linear mixing often raises Recall at the cost of Precision, copy-and-paste methods may hurt Recall on large lesions, and photometric blending favors robustness but can underfit boundaries. Our approach reduces these trade-offs and remains stable, as reflected by smaller standard deviations (four runs).

\begin{table*}[!ht]
    \vspace{-0.4cm}
        \belowrulesep=0pt
        \aboverulesep=0pt
        \caption{Boundary sensitivity of mixing-based methods on Kvasir-SEG, PICCOLO, CVC ClinicDB and NPC-LES.}
        \label{Table:boundary}
    \vspace{-0.2cm}
        \centering
        
    \resizebox{\linewidth}{!}{
        \setlength{\tabcolsep}{0pt}
        \begin{tabular*}{\hsize}{@{}@{\extracolsep{\fill}}c|c|c|c|ccc|ccc|ccc|ccc|}
        \toprule
        \multirow{2}{*}{Dataset} &
        \multirow{2}{*}{Methods} &
        \multirow{2}{*}{$\text{HD}_{95}\downarrow$} &
        \multirow{2}{*}{$\text{ASSD}\downarrow$} &

        \multicolumn{3}{c|}{$\delta=$2px} &
        \multicolumn{3}{c|}{$\delta=$5px} &
        \multicolumn{3}{c|}{$\delta=$10px}&
        \multirow{2}{*}{$\text{BIoU}\uparrow$} \\
        \cmidrule(lr){5-7}\cmidrule(lr){8-10}\cmidrule(lr){11-13}
        & & & &
        $\text{B-P\%}\uparrow$ & $\text{B-R\%}\uparrow$ & $\text{B-F1\%}\uparrow$ & 
        $\text{B-P\%}\uparrow$ & $\text{B-R\%}\uparrow$ & $\text{B-F1\%}\uparrow$ & 
        $\text{B-P\%}\uparrow$ & $\text{B-R\%}\uparrow$ & $\text{B-F1\%}\uparrow$ \\
        \hline
        \multirow{11}{*}{\makecell{\sm{7} Kvasir \\ \sm{7} -SEG}}
        &\sm{7} CutMix~\cite{CutMix} \textit{\fontsize{5}{18}\selectfont (ICCV 2019)}
        &48.32 & 11.72 & 44.21 & 45.08 & 44.18 & 68.28 & 70.16 & 68.48 & 79.11 & 81.78 & 79.56&27.54\\
        &\sm{7} Mixup~\cite{mixup_2018_ICLR} \textit{\fontsize{5}{18}\selectfont (ICLR 2018)}
        &50.65 & 13.35 & 37.19 & 37.63 & 37.03 & 61.71 & 62.23 & 61.36 & 75.74 & 76.68 & 75.48 &22.72\\
        &\sm{7} GridMix~\cite{GridMix}  \textit{\fontsize{5}{18}\selectfont (PR 2021)}
        &65.27 & 17.47 & 23.83 & 32.70 & 26.97 & 42.10 & 57.53 & 47.48 & 55.06 & 73.89 & 61.60&16.16\\
        &\sm{7} SmoothMix~\cite{SmoothMix}  \textit{\fontsize{5}{18}\selectfont (CVPR 2020)}
        &48.15 & 11.36 & 45.82 & 47.36 & 46.12 & 67.95 & 70.76 & 68.67 & 78.09 & 81.89 & 79.18&28.73\\
        &\sm{7} PuzzleMix~\cite{PuzzleMix}  \textit{\fontsize{5}{18}\selectfont (ICML 2020)}
        &44.73 & 11.18 & 40.14 & 40.33 & 39.88 & 66.39 & 67.10 & 66.16 & 79.44 & 80.75 & 79.38&24.76\\
        &\sm{7} AugMix~\cite{AugMix}  \textit{\fontsize{5}{18}\selectfont (ICLR 2020)}
        &43.80 & 11.30 & 39.68 & 38.95 & 39.03 & 65.91 & 65.31 & 65.14 & 79.39 & 79.30 & 78.80&24.30\\
        &\sm{7} PixMix~\cite{PixMix}  \textit{\fontsize{5}{18}\selectfont (CVPR 2022)}
        &44.69 & 11.59 & 42.77 & 41.62 & 41.83 & 67.73 & 66.48 & 66.55 & 79.92 & 79.24 & 78.92&26.07\\
        &\sm{7} DiffuseMix~\cite{DiffuseMix}  \textit{\fontsize{5}{18}\selectfont (CVPR 2024)}
        &45.93 & 11.03 & 43.20 & 42.24 & 42.39 & 68.89 & 67.90 & 67.86 & 81.41 & 81.04 & 80.58&26.36\\
        &\sm{7} HSMix~\cite{HSMix}  \textit{\fontsize{5}{18}\selectfont (InfFus 2025)}
        &48.36 & 11.69 & 43.72 & 44.52 & 43.69 & 67.75 & 69.52 & 67.99 & 78.91 & 81.75 & 79.53&27.21\\
        &\sm{7} \bf{Ours} 
        &\best{40.43} & \best{9.68} & \best{50.13} & \best{49.05} & \best{49.26} & \best{73.26} & \best{72.71} & \best{72.47} & \best{83.38} & \best{83.41} & \best{82.79}&\best{31.38}\\
        \hline
        \multirow{11}{*}{\makecell{\sm{7} PICCOLO}}
        &\sm{7} CutMix~\cite{CutMix} \textit{\sm{6} (ICCV 2019)}
        & 52.07 & 12.96 & 48.33 & 49.65 & 48.65 & 65.19 & 67.30 & 65.76 & 74.68 & 76.92 & 75.16&30.24\\
        &\sm{7} Mixup~\cite{mixup_2018_ICLR}  \textit{\sm{6} (ICLR 2018)}
        & 58.38 & 15.85 & 41.13 & 40.87 & 40.52 & 61.26 & 61.13 & 60.47 & 73.39 & 73.47 & 72.55&24.84\\
        &\sm{7} GridMix~\cite{GridMix}  \textit{\sm{6} (PR 2021)}
        & 83.36 & 22.83 & 18.83 & 31.75 & 22.60 & 31.98 & 52.03 & 37.98 & 43.51 & 67.17 & 50.69&12.69\\
        &\sm{7} SmoothMix~\cite{SmoothMix}  \textit{\sm{6} (CVPR 2020)}
        & 53.52 & 13.12 & 47.27 & 48.80 & 47.70 & 65.09 & 67.36 & 65.63 & 75.42 & 77.90 & 75.92&29.63\\
        &\sm{7} PuzzleMix~\cite{PuzzleMix}  \textit{\sm{6} (ICML 2020)}
        & 54.01 & 14.57 & 44.23 & 44.91 & 44.21 & 62.75 & 63.71 & 62.71 & 73.55 & 74.28 & 73.29&27.33\\
        &\sm{7} AugMix~\cite{AugMix}  \textit{\sm{6} (ICLR 2020)}    
        & 46.98 & 11.35 & 47.83 & 48.76 & 47.90 & 66.58 & 68.67 & 67.01 & 77.32 & 80.02 & 77.97&29.84\\
        &\sm{7} PixMix~\cite{PixMix}  \textit{\sm{6} (CVPR 2022)}
        & 50.23 & 13.15 & 47.13 & 47.59 & 46.95 & 66.20 & 67.51 & 66.25 & 76.48 & 78.11 & 76.51&29.61\\
        &\sm{7} DiffuseMix~\cite{DiffuseMix}  \textit{\sm{6} (CVPR 2024)}
        & 70.60 & 17.78 & 44.99 & 45.26 & 44.50 & 62.50 & 63.33 & 61.96 & 72.48 & 73.21 & 71.57&27.75\\
        &\sm{7} HSMix~\cite{HSMix}  \textit{\sm{6} (InfFus 2025)}
        & 49.03 & 12.55 & 47.85 & 48.32 & 47.76 & 66.55 & 67.31 & 66.39 & 77.34 & 77.74 & 76.70&29.81\\
        &\sm{7} \bf{Ours} 
        & \best{31.46} & \best{6.99} & \best{57.03} & \best{57.45} & \best{57.02} & \best{74.90} & \best{76.03} & \best{75.17} & \best{84.20} & \best{85.53} & \best{84.55}&\best{37.05}\\
        \hline
        \multirow{11}{*}{\makecell{\sm{7} CVC \\ \sm{7} ClinicDB}}
        &\sm{7} CutMix~\cite{CutMix} \textit{\sm{6} (ICCV 2019)}
        &24.25 & 6.27 & 41.13 & 38.93 & 39.89 & 74.04 & 71.46 & 72.46 & 89.41 & 87.83 & 88.26&25.37\\
        &\sm{7} Mixup~\cite{mixup_2018_ICLR}  \textit{\sm{6} (ICLR 2018)}
        &34.41 & 9.22 & 26.03 & 23.84 & 24.78 & 55.22 & 51.33 & 52.95 & 79.09 & 75.04 & 76.64&15.64\\
        &\sm{7} GridMix~\cite{GridMix}  \textit{\sm{6} (PR 2021)}
        &78.84 & 19.50 & 17.69 & 22.32 & 19.40 & 36.52 & 46.00 & 40.04 & 53.72 & 66.12 & 58.28&11.50\\
        &\sm{7} SmoothMix~\cite{SmoothMix}  \textit{\sm{6} (CVPR 2020)}
        &23.48 & 5.89 & 46.15 & 44.21 & 45.05 & 77.13 & 75.19 & 75.94 & 90.09 & 89.17 & 89.39&28.93\\
        &\sm{7} PuzzleMix~\cite{PuzzleMix}  \textit{\sm{6} (ICML 2020)}
        &42.75 & 12.90 & 24.94 & 22.45 & 23.49 & 49.12 & 44.69 & 46.48 & 71.20 & 65.72 & 67.84&14.05\\
        &\sm{7} AugMix~\cite{AugMix}  \textit{\sm{6} (ICLR 2020)}  
        &23.34 & 6.13 & 38.47 & 35.58 & 36.89 & 71.42 & 67.09 & 69.01 & 89.94 & 86.33 & 87.86&23.24\\
        &\sm{7} PixMix~\cite{PixMix}  \textit{\sm{6} (CVPR 2022)}
        &28.67 & 8.05 & 32.49 & 30.55 & 31.35 & 62.68 & 59.86 & 60.93 & 83.58 & 81.25 & 81.96&19.47\\
        &\sm{7} DiffuseMix~\cite{DiffuseMix}  \textit{\sm{6} (CVPR 2024)}
        &27.30 & 7.57 & 36.22 & 33.26 & 34.55 & 67.85 & 62.95 & 65.07 & 86.39 & 82.01 & 83.81&21.71\\
        &\sm{7} HSMix~\cite{HSMix}  \textit{\sm{6} (InfFus 2025)}
        &33.17 & 8.75 & 30.69 & 29.14 & 29.74 & 61.11 & 58.58 & 59.52 & 82.22 & 80.17 & 80.77&18.67\\
        &\sm{7} \bf{Ours} 
        &\best{20.46} & \best{5.11} & \best{50.57} & \best{47.98} & \best{49.12} & \best{81.93} & \best{79.25} & \best{80.35} & \best{92.69} & \best{90.69} & \best{91.42}&\best{31.56}\\

        \hline
        
        \multirow{10}{*}{\makecell{\sm{7} NPC\\ \sm{7} -LES}}
        &\sm{7} CutMix~\cite{CutMix} \textit{\sm{6} (ICCV 2019)}
            &59.51 & 16.18 & 28.46 & 29.66 & 28.95 & 50.00 & 52.16 & 50.89 & 64.13 & 66.93 & 65.27&17.30\\
        &\sm{7} Mixup~\cite{mixup_2018_ICLR}  \textit{\sm{6} (ICLR 2018)}
            &56.61 & 14.95 & 29.01 & 29.78 & 29.30 & 51.05 & 52.48 & 51.60 & 66.34 & 68.29 & 67.08&17.51\\
        &\sm{7} GridMix~\cite{GridMix}  \textit{\sm{6} (PR 2021)}
            &72.34 & 19.80 & 16.25 & 25.62 & 19.35 & 30.80 & 48.55 & 36.67 & 43.24 & 67.36 & 51.28&11.28\\
        &\sm{7} SmoothMix~\cite{SmoothMix}  \textit{\sm{6} (CVPR 2020)}
            &56.55 & 15.48 & 28.84 & 30.55 & 29.56 & 50.51 & 53.50 & 51.76 & 64.33 & 67.94 & 65.83&17.71\\
        &\sm{7} PuzzleMix~\cite{PuzzleMix}  \textit{\sm{6} (ICML 2020)}
            &58.45 & 16.33 & 28.51 & 29.73 & 29.02 & 50.43 & 52.50 & 51.27 & 64.22 & 66.55 & 65.14&17.40\\
        &\sm{7} AugMix~\cite{AugMix}  \textit{\sm{6} (ICLR 2020)}    
            &53.49 & 14.55 & 29.42 & 29.88 & 29.57 & 51.61 & 52.57 & 51.92 & 66.61 & 67.93 & 67.06&17.68\\
        &\sm{7} PixMix~\cite{PixMix}  \textit{\sm{6} (CVPR 2022)}
            &55.49 & 15.03 & 29.70 & 30.46 & 29.99 & 51.90 & 53.35 & 52.46 & 66.33 & 68.24 & 67.07&18.00\\
        &\sm{7} DiffuseMix~\cite{DiffuseMix}  \textit{\sm{6} (CVPR 2024)}
            &54.14 & 14.54 & 29.77 & 30.86 & 30.19 & 52.15 & 54.23 & 52.96 & 66.46 & 69.21 & 67.53&18.09\\
        &\sm{7} HSMix~\cite{HSMix}  \textit{\sm{6} (InfFus 2025)}
            &58.15 & 15.43 & 28.99 & 29.99 & 29.37 & 50.68 & 52.58 & 51.42 & 65.31 & 67.69 & 66.21&17.53\\
        &\sm{7} \bf{Ours} 
            &\best{45.68} & \best{12.23} & \best{34.23} & \best{34.50} & \best{34.28} & \best{58.34} & \best{59.07} & \best{58.55} & \best{72.52} & \best{73.57} & \best{72.84}&\best{20.76}\\
        \bottomrule
        \end{tabular*}
        \vspace{-0.5cm}
    }
    \vspace{-0.6cm}
    \end{table*} 

\subsubsection{Qualitative analysis}
Visual comparisons on Kvasir-SEG, PICCOLO, CVC-ClinicDB, and NPC-LES (Figs.~\ref{fig:Different_methods_Kvasir_CVC}) show that our method finds the lesion location more accurately under difficult conditions, including gradual lesion-mucosa changes, specular highlights, mucus streaks, motion blur, and uneven colors. With better localization, the predicted masks also match the real boundaries more closely: small or thin lesions are kept, false holes are reduced, and over-segmented areas are fewer. On NPC-LES, which has many early lesions and site-specific backgrounds, our results stay consistent in ambiguous regions and keep the correct shape. Overall, same-mask appearance mixing improves lesion localization without label ambiguity, and, together with our training pipeline, keeps realistic textures while delivering precise boundaries.





\begin{table}[!t]
    \vspace{-0.2cm}
    \belowrulesep=0pt
    \aboverulesep=0pt
    \caption{Comparison with Generative dataset augmentation methods on NPC-LES (Mean \text{\sm{6}$\pm$ Standard deviation}, n=4).}
    \label{Table:different_framework__NPC_LES_Diffusion_based}
    \centering
    \setlength{\tabcolsep}{1pt}
    \begin{tabular*}{\hsize}{@{}@{\extracolsep{\fill}}c|cccc@{}}
        \toprule
        Metric &
        ControlPolypNet~\cite{ControlPolypNet}  &
        GenSRRFI~\cite{GenSRRFI}  &
        SatSynth~\cite{SatSynth}  &
        Ours \\
        \hline
        $\text{mIoU}$       & 88.96\sm{6}$\pm$0.28
                            & {89.21}\sm{6}$\pm$0.33
                            & 78.53\sm{6}$\pm$1.29
                            & \best{90.10}\sm{6}$\pm$0.66 \\
        $\text{PA}$         & 95.07\sm{6}$\pm$0.20
                            & {95.20}\sm{6}$\pm$0.54
                            & 82.46\sm{6}$\pm$1.96
                            & \best{95.49}\sm{6}$\pm$0.41 \\
        $\text{Recall}$     & 91.94\sm{6}$\pm$0.49
                            & {92.01}\sm{6}$\pm$0.39
                            & 83.94\sm{6}$\pm$2.01
                            & \best{92.78}\sm{6}$\pm$1.20 \\
        $\text{Precision}$  & {93.56}\sm{6}$\pm$0.52
                            & 93.55\sm{6}$\pm$0.45
                            & 83.92\sm{6}$\pm$1.63
                            & \best{94.49}\sm{6}$\pm$0.83 \\
        $\text{DSC}$       & 91.44\sm{6}$\pm$0.30
                            & {91.58}\sm{6}$\pm$0.40
                            & 78.21\sm{6}$\pm$2.22
                            & \best{92.57}\sm{6}$\pm$0.37 \\
        \bottomrule
    \end{tabular*}
    \vspace{-0.8cm}
\end{table}

\subsubsection{Boundary prediction performance}
Across all four datasets shown in Table \ref{Table:boundary}, our method consistently excels on boundary-sensitive metrics: lower $\text{HD}_{95}$ and $\text{ASSD}$, higher $\text{B-F1}$ at both strict and moderate tolerances, and the best $\text{BIoU}$. Notably, improvements in $\text{B-F1}$ persist even at small $\delta$, indicating robustness on thin and tortuous contours, while at larger $\delta$ we observe concurrent gains in $\text{B-P}$ and $\text{B-R}$, suggesting reduced over- and under-segmentation. The overall increase in $\text{BIoU}$ over the strongest mixing baseline further evidences more accurate overlap along fine boundaries.
These trends align tightly with our core design. 
Mask-Consistent Paired Mixing constrains appearance mixing within a shared semantic mask, explicitly avoiding cross-image misalignment and soft label blurring at lesion edges, which lowers boundary distances and boosts $\text{B-F1}$. Real-Anchored Learnable Annealing then learns to down-weight synthetic samples over time so that features re-anchor to the real domain, reinforcing the balance between $\text{B-P}$ and $\text{B-R}$ and stabilizing gains in $\text{BIoU}$. In combination, decreasing boundary distances and increasing boundary agreement occur jointly, reflecting the synergy between mask-consistent mixing and learnable re-anchoring for boundary quality.

\subsubsection{Comparison with Generative dataset augmentation}\label{subsec:Generative}
As reported in Table~\ref{Table:different_framework__NPC_LES_Diffusion_based}, under the same settings our method leads on all five core metrics. Relative to GenSRRFI, the improvements are 0.89\% in $\text{mIoU}$, 0.29\% in $\text{PA}$, 0.77\% in $\text{Recall}$, 0.94\% in $\text{Precision}$, and 0.99\% in $\text{DSC}$. The limited overlap between the $\text{DSC}$ mean–std ranges may relate to enhanced stability in boundary learning. Compared with ControlPolypNet and GenSRRFI, concurrent gains in $\text{Precision}$ and $\text{Recall}$ suggest better control of leakage and misses near complex borders. SatSynth attains lower scores with higher variance, possibly influenced by domain statistics. Qualitatively, we often observe more continuous contours and clearer confidence maps under blurred boundaries, reflective occlusions, and small lesions. Overall, geometry-consistent paired mixing and learnable annealing may jointly contribute to these trends.

\begin{table*}[!t]
    \belowrulesep=0pt
    \aboverulesep=0pt
    \centering
    \caption{Ablation Experiment on Kvasir-SEG, PICCOLO, CVC ClinicDB and NPC-LES (Mean \text{\sm{6}$\pm$Standard deviation}, n=4).}
    \label{Table:ablation_experiment}
    \begin{tabular*}{\hsize}{@{}@{\extracolsep{\fill}}c|ccc|ccccc}
    \toprule
    \multirow{2}{*}{Dataset} &
    \multicolumn{3}{c|}{Ablation Settings} &
    \multirow{2}{*}{$\text{mIoU}$ $ \left( \% \right) $} &  
    \multirow{2}{*}{$\text{PA}$ $ \left( \% \right) $} &  
    \multirow{2}{*}{$\text{Recall}$ $ \left( \% \right) $} & 
    \multirow{2}{*}{$\text{Precision}$ $ \left( \% \right) $} & 
    \multirow{2}{*}{ $\text{DSC}$ $ \left( \% \right) $}
    \\
     \cmidrule(lr){2-4}  
      & Full-Supervised & +MCPMix & +RLA  \\
    \hline
    \multirow{3}{*}{\makecell{Kvasir-SEG \\(Public Dataset)}}     
    & \checkmark&\usym{1F5F4}&\usym{1F5F4}   &84.25\sm{6}$\pm$0.39  &93.21\sm{6}$\pm$0.51  &84.44\sm{6}$\pm$0.51  &89.39\sm{6}$\pm$0.52  &85.55\sm{6}$\pm$0.43 \\ 
    & \checkmark &\checkmark&  \usym{1F5F4}  &88.21\sm{6}$\pm$0.76&96.34\sm{6}$\pm$0.51&87.15\sm{6}$\pm$1.10&92.63\sm{6}$\pm$0.93&87.73\sm{6}$\pm$1.41\\
    & \checkmark &\checkmark&  \checkmark  &\best{88.72}\sm{6}$\pm$0.30&\best{96.55}\sm{6}$\pm$0.75&\best{87.18}\sm{6}$\pm$0.66&\best{93.21}\sm{6}$\pm$0.77&\best{88.13}\sm{6}$\pm$0.33\\
    \hline
    \multirow{3}{*}{\makecell{PICCOLO \\(Public Dataset)}}     
    & \checkmark&\usym{1F5F4}&\usym{1F5F4}   &76.53\sm{6}$\pm$0.92  &90.28\sm{6}$\pm$0.68  &70.22\sm{6}$\pm$1.01  &84.44\sm{6}$\pm$1.25  &72.00\sm{6}$\pm$0.80  \\ 
    & \checkmark &\checkmark&  \usym{1F5F4}  &86.63\sm{6}$\pm$0.78&97.16\sm{6}$\pm$0.68&\best{82.78}\sm{6}$\pm$0.72&89.74\sm{6}$\pm$0.39&\best{85.30}\sm{6}$\pm$0.89\\
    & \checkmark &\checkmark&  \checkmark  &\best{87.11}\sm{6}$\pm$0.59&\best{97.62}\sm{6}$\pm$0.34&81.27\sm{6}$\pm$0.78&\best{92.94}\sm{6}$\pm$0.42&84.24\sm{6}$\pm$0.49\\
    \hline
    \multirow{3}{*}{\makecell{CVC ClinicDB \\(Public Dataset)}}     
    & \checkmark&\usym{1F5F4}&\usym{1F5F4}   &85.33\sm{6}$\pm$0.95  &95.01\sm{6}$\pm$0.88  &74.56\sm{6}$\pm$1.10  & 89.98\sm{6}$\pm$1.00  &84.12\sm{6}$\pm$0.75   \\ 
    & \checkmark &\checkmark&  \usym{1F5F4}   &91.68\sm{6}$\pm$1.23&98.73\sm{6}$\pm$0.21&88.99\sm{6}$\pm$1.69&94.71\sm{6}$\pm$0.84&91.03\sm{6}$\pm$0.44\\
    & \checkmark &\checkmark&  \checkmark  &\best{92.63}\sm{6}$\pm$0.36&\best{98.95}\sm{6}$\pm$0.05&\best{90.16}\sm{6}$\pm$0.92&\best{95.32}\sm{6}$\pm$0.71&\best{91.94}\sm{6}$\pm$0.60\\
    \hline
    \multirow{3}{*}{\makecell{NPC-LES \\(Private Dataset)}}     
    & \checkmark&\usym{1F5F4}&\usym{1F5F4} &84.51\sm{6}$\pm$0.24 &91.82\sm{6}$\pm$0.54 &88.66\sm{6}$\pm$1.02 &89.26\sm{6}$\pm$1.00 &86.67\sm{6}$\pm$0.87 \\ 
    & \checkmark &\checkmark&  \usym{1F5F4}  &89.20\sm{6}$\pm$0.79&95.00\sm{6}$\pm$0.43&92.29\sm{6}$\pm$0.77&93.97\sm{6}$\pm$0.88&91.88\sm{6}$\pm$0.86\\
    & \checkmark &\checkmark&  \checkmark  &\best{90.10}\sm{6}$\pm$0.66&\best{95.49}\sm{6}$\pm$0.41&\best{92.78}\sm{6}$\pm$1.20&\best{94.49}\sm{6}$\pm$0.83&\best{92.57}\sm{6}$\pm$0.37\\

    \bottomrule
    \end{tabular*}
    \vspace{-0.5cm}
    \end{table*}

\subsection{Ablation analysis}
We evaluate three configurations under identical training settings and hyperparameters on Kvasir-SEG, PICCOLO, CVC-ClinicDB, and NPC-LES:
Full-Sup, +MCPMix, and +MCPMix+RLA.
Metrics include $\text{mIoU}$, PA, $\text{Precision}$, $\text{Recall}$, and $\text{DSC}$.
Results are reported in Table~\ref{Table:ablation_experiment}.
Compared with Full-Sup, +MCPMix consistently improves $\text{mIoU}$, $\text{DSC}$, and $\text{Precision}$.
Same-mask linear mixing enlarges the appearance neighborhood without semantic ambiguity and injects diversity.
We observe gains of $\text{mIoU}$ $+4.69\%$, $\text{DSC}$ $+5.21\%$, and $\text{Precision}$ $+4.71\%$.
Higher $\text{Precision}$ indicates fewer false positives in complex backgrounds, and higher ${DSC}$ indicates sharper and more consistent boundaries.
With RLA, $\text{Recall}$, $\text{Precision}$, and PA further increase, and $\text{mIoU}$ and $\text{DSC}$ reach the best performance.
Learnable mixing and loss weighting guided by mild temporal priors and distribution signals shift training from early \enquote{appearance exploration} to late \enquote{real convergence}.
Pixel accuracy and $\text{Precision}$ rise, residual synthetic bias shrinks, and $\text{Recall}$ improves slightly without harming $\text{Precision}$, which yields a better $\text{Recall}$--$\text{Precision}$ balance.
Additional gains in $\text{mIoU}$ and $\text{DSC}$ confirm the consolidation of regional consistency and boundary refinement.
In summary, MCPMix introduces ambiguity-free appearance diversity that enhances invariance and robustness, while RLA mitigates synthetic-domain overfitting by re-anchoring to the real distribution. Together they form an expansion-then-alignment loop and yield a balanced performance across all metrics.

\begin{table}[!t]
    \vspace{-0.2cm}
    \centering
    \caption{Comparison with mixing-based methods on ISIC 2017 dataset(Mean \text{\sm{6}$\pm$ Standard deviation}, n=4).}
    \setlength{\tabcolsep}{2pt}
    \resizebox{\linewidth}{!}{
    \begin{tabular}{ccccccc}
        \toprule
        Method   & $\text{mIoU}$           & $\text{PA}$   & $\text{Recall}$        &      $\text{Precision}$ &       $\text{DSC}$   \\
        \hline
        CutMix \textit{\sm{6} (ICCV 2019)}
             &81.32\sm{6}$\pm$1.26
             &92.46\sm{6}$\pm$0.88
             &76.70\sm{6}$\pm$1.69
             &95.13\sm{6}$\pm$1.01
             &81.83\sm{6}$\pm$1.54\\
        Mixup \textit{\sm{6} (ICLR 2018)}
             &80.48\sm{6}$\pm$1.38
             &92.48\sm{6}$\pm$0.86
             &74.74\sm{6}$\pm$2.31
             &95.84\sm{6}$\pm$0.90
             &80.66\sm{6}$\pm$1.69\\
        GridMix \textit{\sm{6} (PR 2021)}
             &81.24\sm{6}$\pm$1.33
             &92.59\sm{6}$\pm$0.82
             &76.49\sm{6}$\pm$1.79
             &94.84\sm{6}$\pm$1.05
             &81.88\sm{6}$\pm$1.60\\
        SmoothMix \textit{\sm{6} (CVPR 2020)}
             &81.20\sm{6}$\pm$1.12
             &92.67\sm{6}$\pm$0.85
             &{78.04}\sm{6}$\pm$1.44
             &93.62\sm{6}$\pm$1.06
             &81.39\sm{6}$\pm$1.38\\
        PuzzleMix \textit{\sm{6} (ICML 2020)}
             &80.05\sm{6}$\pm$1.54
             &91.77\sm{6}$\pm$1.04
             &73.78\sm{6}$\pm$2.28
             &95.98\sm{6}$\pm$0.93
             &80.19\sm{6}$\pm$1.84\\
        AugMix \textit{\sm{6} (ICLR 2020)}
             &{82.16}\sm{6}$\pm$1.19
             &{92.98}\sm{6}$\pm$0.80
             &77.79\sm{6}$\pm$1.56
             &95.53\sm{6}$\pm$1.03
             &{83.13}\sm{6}$\pm$1.28\\
        PixMix \textit{\sm{6} (CVPR 2022)}
             &81.35\sm{6}$\pm$1.19
             &92.64\sm{6}$\pm$0.82
             &75.75\sm{6}$\pm$1.66
             &\best{96.33}\sm{6}$\pm$0.70
             &81.86\sm{6}$\pm$1.41\\
        DiffuseMix \textit{\sm{6} (CVPR 2024)}
             &81.66\sm{6}$\pm$0.85
             &92.65\sm{6}$\pm$0.27
             &76.39\sm{6}$\pm$1.58
             &{96.19}\sm{6}$\pm$0.42
             &82.46\sm{6}$\pm$1.12\\
        HSMix \textit{\sm{6} (InfFus 2025)}
             &80.85\sm{6}$\pm$0.96
             &92.50\sm{6}$\pm$0.96
             &75.44\sm{6}$\pm$1.45
             &95.62\sm{6}$\pm$0.54
             &80.94\sm{6}$\pm$1.22\\
        \bf{Ours}
             &\best{83.13}\sm{6}$\pm$0.84
             &\best{93.55}\sm{6}$\pm$0.60
             &\best{79.84}\sm{6}$\pm$0.82
             &94.91\sm{6}$\pm$0.23
             &\best{83.93}\sm{6}$\pm$0.86\\
        \bottomrule
    \end{tabular}
    }
    \label{Table:ISIC}
    \vspace{-0.4 cm}
\end{table}

\subsubsection{Generative experiment out of distribution}\label{subsec:ISIC}
\begin{figure}[t]
    \centering
    \vspace{-0.2cm}
    \includegraphics[width=\columnwidth ]{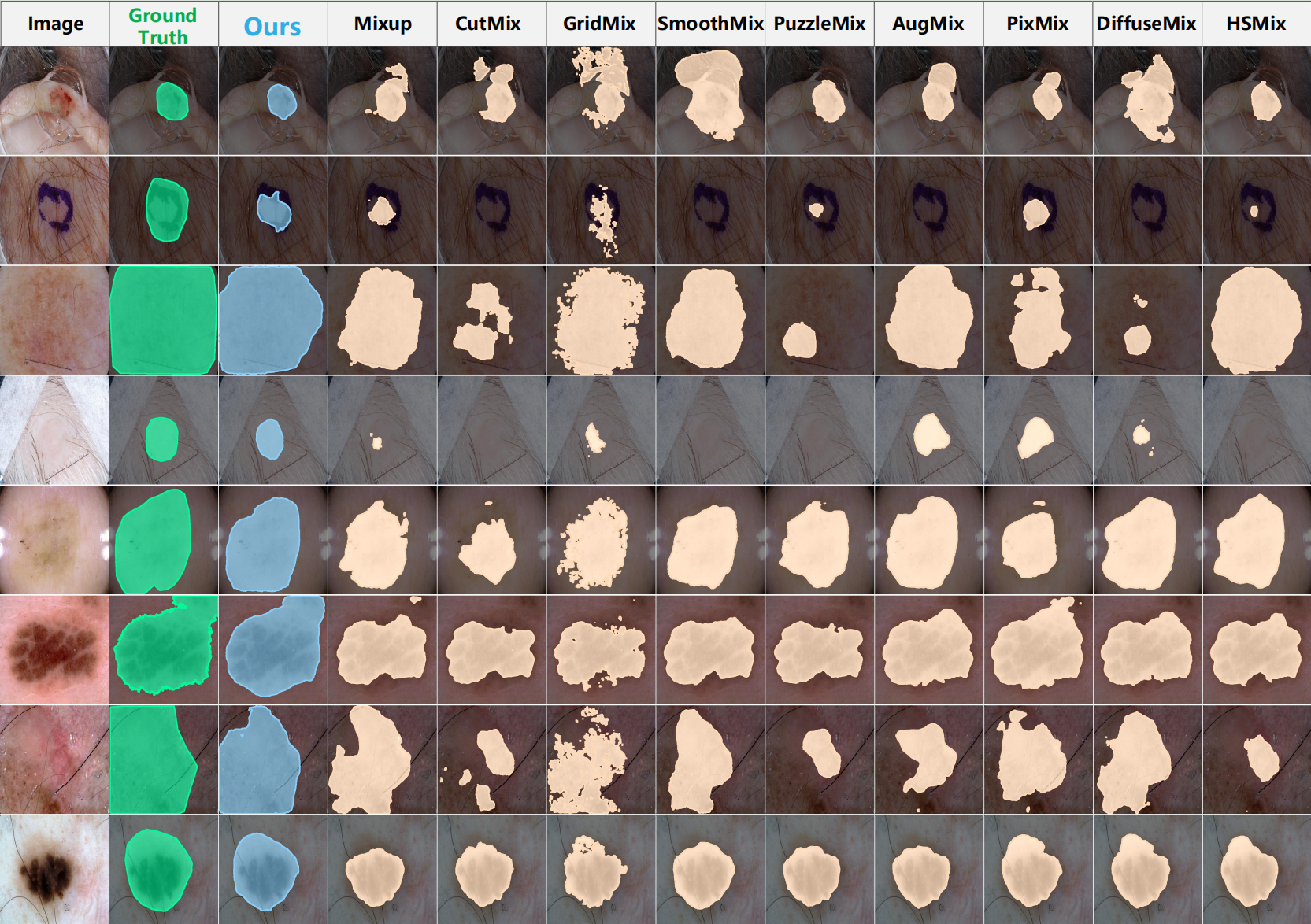} 
    \caption{Visualization of different methods on \textit{ISIC 2017} dataset.}
    \label{fig:Different_methods_ISIC}
    \vspace{-0.5 cm}
\end{figure}

We evaluate cross-domain generalization on the ISIC 2017 dermoscopic dataset, with results summarized in Table~\ref{Table:ISIC}.
Our method achieves the best $\text{mIoU}$, $\text{DSC}$, $\text{Recall}$, and $\text{PA}$, indicating higher overall quality, better lesion coverage, and more consistent boundaries.
Concurrent gains in $\text{mIoU}$ and $\text{DSC}$ reflect a balanced overlap between precision and recall, which is relevant for clinical use.
Mask-consistent paired mixing with hard label supervision together with learnable annealing reduces synthetic-real shift and improves robustness under blurred edges and low contrast.
As shown in Fig.~\ref{fig:Different_methods_ISIC}, faint or partially occluded rims are delineated more reliably with less leakage into background.

\begin{table}[!h]
    \belowrulesep=0pt
    \aboverulesep=0pt
    \centering
    \vspace{-0.4cm}
    \caption{Comparisons with different backbones on NPC-LES(Mean \text{\sm{6}$\pm$ Standard deviation}, n=4).}
    \setlength{\tabcolsep}{2pt}
    \vspace{-0.1cm}
    \resizebox{\linewidth}{!}{
        \begin{tabular}{c|ccccc}
            \toprule
            Method   & $\text{mIoU}$ & $\text{PA}$ & $\text{Recall}$ & $\text{Precision}$ &  $\text{DSC}$ \\
    \hline
    \sm{7}SegFormer
      & \sm{7}85.51\sm{5}$\pm$0.87
      & \sm{7}93.82\sm{5}$\pm$0.78
      & \sm{7}90.66\sm{5}$\pm$0.65
      & \sm{7}89.26\sm{5}$\pm$0.79
      & \sm{7}86.67\sm{5}$\pm$0.38 \\
    \sm{7}SegFormer $\mathbf{+}$ Ours
      & \sm{7}{90.10}\sm{5}$\pm$0.66
      & \sm{7}{95.49}\sm{5}$\pm$0.41
      & \sm{7}{92.78}\sm{5}$\pm$1.20
      & \sm{7}{94.49}\sm{5}$\pm$0.83
      & \sm{7}{92.57}\sm{5}$\pm$0.37 \\
    \hline
    \sm{7}SegNet
      & \sm{7}81.64\sm{5}$\pm$0.37
      & \sm{7}93.78\sm{5}$\pm$0.40
      & \sm{7}90.87\sm{5}$\pm$0.60
      & \sm{7}81.16\sm{5}$\pm$0.47
      & \sm{7}79.53\sm{5}$\pm$0.89 \\
    \sm{7}SegNet \& Ours
      & \sm{7}84.60\sm{5}$\pm$0.91
      & \sm{7}94.78\sm{5}$\pm$0.18
      & \sm{7}91.94\sm{5}$\pm$0.45
      & \sm{7}86.11\sm{5}$\pm$0.72
      & \sm{7}84.57\sm{5}$\pm$0.33 \\
    \hline
    \sm{7}DeepLabV3+
      & \sm{7}84.35\sm{5}$\pm$0.40
      & \sm{7}93.38\sm{5}$\pm$0.93
      & \sm{7}87.83\sm{5}$\pm$0.78
      & \sm{7}90.91\sm{5}$\pm$1.02
      & \sm{7}86.01\sm{5}$\pm$0.84 \\
    \sm{7}DeepLabV3+ \& Ours
      & \sm{7}87.48\sm{5}$\pm$0.43
      & \sm{7}94.73\sm{5}$\pm$0.28
      & \sm{7}93.12\sm{5}$\pm$0.81
      & \sm{7}89.99\sm{5}$\pm$0.83
      & \sm{7}88.75\sm{5}$\pm$1.33 \\
    \hline
    \sm{7}UNet
      & \sm{7}74.04\sm{5}$\pm$1.01
      & \sm{7}91.52\sm{5}$\pm$0.37
      & \sm{7}89.06\sm{5}$\pm$1.06
      & \sm{7}69.72\sm{5}$\pm$0.97
      & \sm{7}68.35\sm{5}$\pm$1.25 \\
    \sm{7}UNet \& Ours
      & \sm{7}77.72\sm{5}$\pm$1.19
      & \sm{7}93.49\sm{5}$\pm$0.41
      & \sm{7}91.70\sm{5}$\pm$0.80
      & \sm{7}73.81\sm{5}$\pm$1.20
      & \sm{7}72.95\sm{5}$\pm$0.89 \\
    \bottomrule
        \end{tabular}
    }
    \label{Table:different_backbone}
    \vspace{-0.7 cm}
\end{table}

\subsection{Comparisons with different backbones}
Across various backbone networks, our method consistently enhances segmentation performance, demonstrating strong model-agnostic properties (Table~\ref{Table:different_backbone}). It effectively improves both CNN-based backbones (e.g., DeepLabV3+, UNet, SegNet) and Transformer-based backbones (e.g., SegFormer), indicating that its benefits are not tied to a specific model design. The improvements arise from data- and feature-level regularization rather than structural dependence, suggesting good transferability and generality. Overall, the method serves as a plug-and-play enhancement module that yields stable and consistent gains across diverse segmentation frameworks.

\section{Discussion}\label{sec_discussion}
This section studies how MCPMix and RLA shape representations and optimization, evaluates learnable versus fixed mixing, and explains why classical inter sample mixing destabilizes boundary gradients through distribution dynamics, schedule comparisons, and a gradient level analysis.

\subsection{Distribution Analysis}

\begin{figure}[h]
    \centering
    \vspace{-0.3cm}
    \includegraphics[width=8cm ]{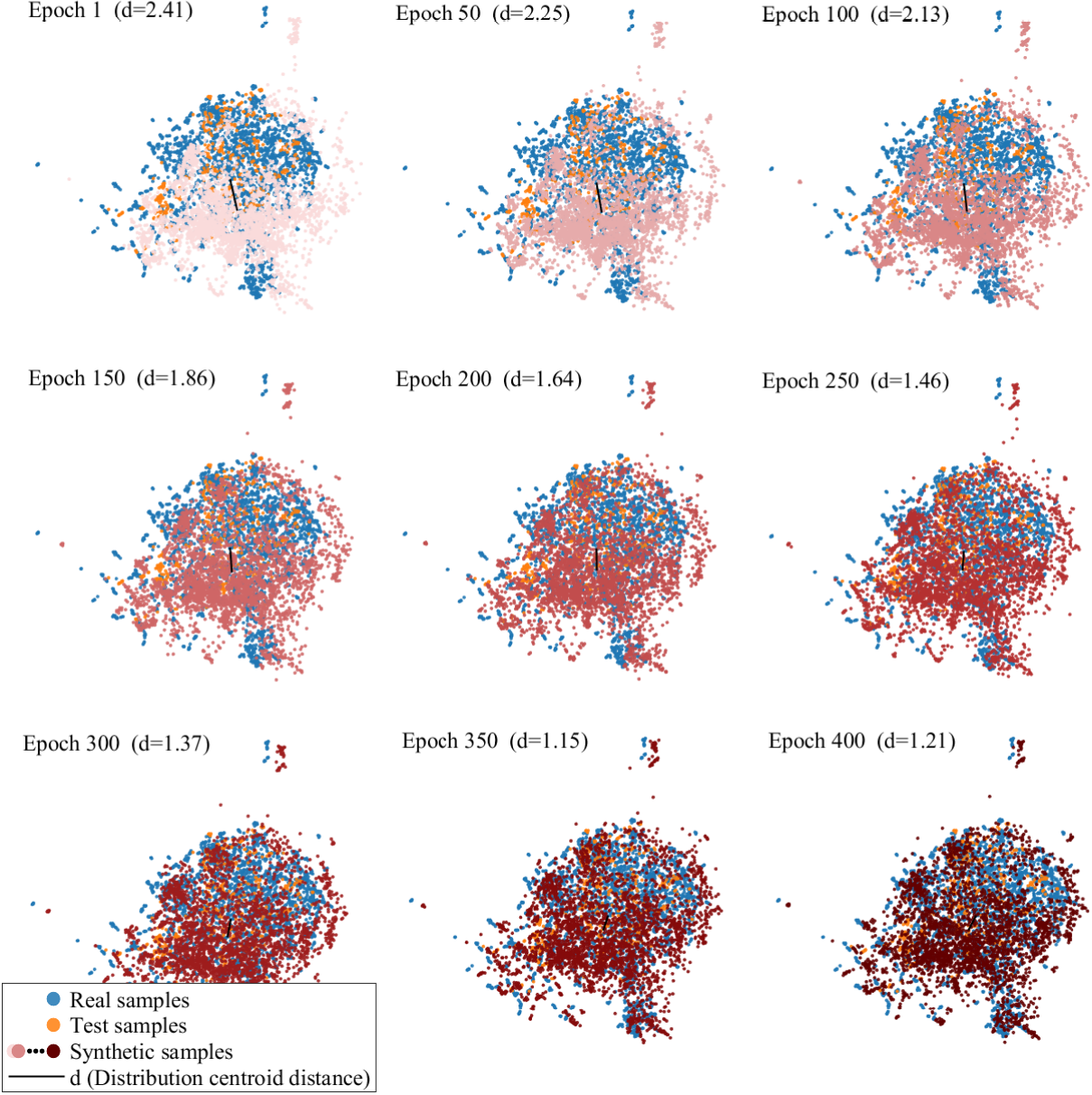} 
    \caption{Distribution changes during training process.}
    \label{fig:distribution}
    \vspace{-0.6 cm}
\end{figure}

We first quantify how the representation of $I_{\text{mix}}$ evolves during training. Specifically, we embed features of mixed and real samples with a fixed backbone (ResNet-50, IMAGENET1K\_V2; 2048-d) and project them to two dimensions via UMAP at nine checkpoints over 400 epochs. We then measure cross-set separation as the Euclidean distance between the corresponding centroids. Empirically, the trajectories exhibit a larger early separation followed by a monotonic downward trend in centroid distance (Fig.~\ref{fig:distribution}). This pattern is consistent with RLA gradually re-anchoring mixed samples toward the real domain and, consequently, reducing learning deviation. However, we emphasize that this analysis queries a fixed ImageNet feature space. Therefore, it captures relative drift rather than full distributional discrepancies.

\subsection{Trainable RLA vs. handcrafted mixing schedules}
\begin{figure}[h]
    \centering 
    \includegraphics[width=\columnwidth ]{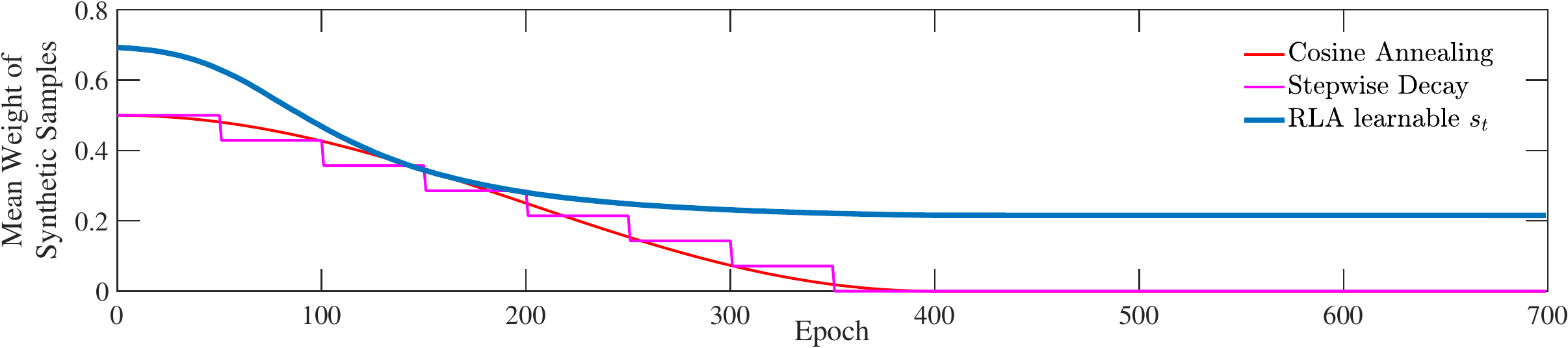} 
    \caption{Different mixing weight strategies.}
    \label{fig:RLA_curve}
    \vspace{-0.6 cm}
\end{figure}

RLA uses a cosine-annealed prior for $s_t$ that decays to near zero by epoch 400. The learned controller is data-driven, so $s_t$ is not forced to follow this prior exactly.
It adaptively adjusts to the learning dynamics. In a representative run (Fig.~\ref{fig:RLA_curve}), $s_t$ receives downward gradients earlier than the preset curve and ultimately converges around a more suitable in that experiment rather than the preset prior value 0.
To further assess the benefit of RLA, we compare it with two fixed mixing schedules, summarized in Table~\ref{Table:RLA}. The first is a stepwise decay, where the weight of $I_s$ is initially sampled from $[0,1]$, then reduced every 50 epochs after epoch 100, and clipped to 0 at epoch 400. The second is a cosine schedule, $\lambda_{cos}=r\cdot 0.25\bigl(1+\cos(\pi t/400)\bigr)$ with $r\in[0,1]$ and $t\in[0,400]$. The comparisons indicate that RLA tends to converge to a more suitable, data-driven value than the fixed priors, thereby improving performance and reducing reliance on handcrafted schedules.
\begin{table}[!h]
    \centering
    \vspace{-0.4cm}
    \caption{Validation of different mixing weight strategies on NPC-LES.}
    \setlength{\tabcolsep}{2pt}
    \vspace{-0.1cm}
    \resizebox{\linewidth}{!}{
        \begin{tabular}{cccccc}
            \toprule
            Method   & $\text{mIoU}$ & $\text{PA}$ & $\text{Recall}$ & $\text{Precision}$ &  $\text{DSC}$ \\
            \hline
            Stepwise Decay
              & {89.60}\sm{6}$\pm$0.72
              & {95.34}\sm{6}$\pm$0.39
              & {92.22}\sm{6}$\pm$0.94
              & {94.27}\sm{6}$\pm$0.58
              & {91.79}\sm{6}$\pm$0.44 \\
            Cosine Annealing
              & 89.45\sm{6}$\pm$0.34
              & 94.09\sm{6}$\pm$0.32
              & 92.19\sm{6}$\pm$0.62
              & 93.88\sm{6}$\pm$0.99
              & 91.68\sm{6}$\pm$0.25 \\
            Ours
              & \best{90.10}\sm{6}$\pm$0.66
              & \best{95.49}\sm{6}$\pm$0.41
              & \best{92.78}\sm{6}$\pm$1.20
              & \best{94.49}\sm{6}$\pm$0.83
              & \best{92.57}\sm{6}$\pm$0.37 \\
            \bottomrule
        \end{tabular}
    }
    \label{Table:RLA}
    \vspace{-0.3cm}
\end{table}

\subsection{Why classical mixing causes unstable gradients}\label{subsec:why_gradient}
Classical inter sample mixing forms $I'=\lambda I_a+(1-\lambda)I_b$ with a soft label $\tilde{M}=\lambda M_a+(1-\lambda)M_b$.
Since $M_a$ and $M_b$ are rarely pixel aligned, boundary pixels often satisfy $\tilde{M}\approx0.5$.
Under cross entropy the per pixel update scales with $\hat{M}-\text{label}$, and under soft DSC fractional labels reduce contributions, so the contour signal is weakened.
Across mini batches different partners and $\lambda$ values make $\tilde{M}$ at the same coordinates drift within $(0,1)$, which flips the update direction between foreground and background and yields high variance low mean gradients that favor a wide transition band instead of a sharp edge.
In contrast, our same mask appearance mixing (Eq.~(\ref{eq:mixup})) builds $I_{\mathrm{mix}}=(1-s_t)I_r+s_t I_s$ while keeping the hard label $Y_{\mathrm{mix}}\equiv M$.
Every pixel of $I_{\mathrm{mix}}$ is consistent with $M$, preserving strong and stable contour gradients while appearance diversity is injected through $I_s$.

\section{Conclusion}\label{sec_Conclusion}
We introduce a paired, diffusion-guided augmentation for endoscopic segmentation. MCPMix provides label-preserving diversity under fixed geometry, while an adaptive reanchoring scheme progressively restores real-domain dominance, giving a smooth path from synthetic to real. Departing from conventional mixing and generate-and-append usage, the framework mitigates domain drift and sharpens boundaries, showing consistent gains across experiments on multiple public and private clinical datasets.


\bibliographystyle{IEEEtran}
\bibliography{ref}

\vfill

\end{document}